\documentclass[9.5pt,journal,compsoc]{IEEEtran}

\usepackage{xcolor,soul,framed} 

\colorlet{shadecolor}{yellow}
\usepackage[pdftex]{graphicx}
\graphicspath{{../pdf/}{../jpeg/}}
\DeclareGraphicsExtensions{.pdf,.jpeg,.png}

\usepackage[cmex10]{amsmath}
\usepackage{array}
\usepackage{mdwmath}
\usepackage{mdwtab}
\usepackage{eqparbox}
\usepackage{url}
\usepackage{amsmath}
\usepackage{amsfonts}
\usepackage[ruled]{algorithm2e}
\usepackage{setspace}
\usepackage{algpseudocode}
\usepackage{multirow}
\usepackage{footnote}
\usepackage{bm}
\usepackage{makecell}
\usepackage{pdflscape}
\usepackage{afterpage}
\usepackage{bbding}
\usepackage{lscape}
\usepackage{booktabs}
\usepackage{tikz}
\usepackage{tabularx}
\usepackage{mathrsfs}
\usepackage[figuresright]{rotating}
\usepackage{graphicx}
\usepackage{subcaption} 

\usepackage{pgfplots}
\pgfplotsset{compat=1.12}
    \pgfplotsset{
        layers/my layer set/.define layer set={
            background,
            main,
            foreground
        }{ },
        set layers=my layer set,
    }

\usepackage[square,sort,comma,numbers]{natbib} 
\usepackage[top=0.4in,bottom=0.4in,left=0.4in,textwidth=7.7in]{geometry}

\usepackage[pagebackref=false,breaklinks=false,linkcolor=red,anchorcolor=blue, citecolor=green,colorlinks,bookmarks=true]{hyperref}
\usepackage{amsfonts}

\usepackage{pifont}

\begin{document}

\title{A Forward and Backward Compatible Framework for Few-shot Class-incremental Pill Recognition}

\author{\IEEEauthorblockN{Jinghua Zhang\thanks{Jinghua Zhang (zhangjingh@foxmail.com), Kai Gao (gaokai\_14@163.com), and Dewen Hu (dwhu@nudt.edu.cn) 
are with the College of Intelligence Science and Technology, National University of Defense Technology (NUDT), Changsha, China. Li Liu (dreamliu2010@gmail.com) is with the College of Electronic Science and Technology, NUDT, Changsha, China.}},
Li Liu,  
Kai Gao,
Dewen Hu
\thanks{Corresponding authors: Dewen Hu and Li Liu}
\thanks{This work was partially supported by the National Key Research and Development Program of China No.2021YFB3100800, the Academy of Finland under grant 331883, the National Natural Science Foundation of China under grant 62376283, and the Key Stone grant (JS2023-03) of the NUDT.
}
}


\IEEEtitleabstractindextext{%
\begin{abstract}
Automatic Pill Recognition (APR) systems are crucial for enhancing hospital efficiency, assisting visually impaired individuals, and preventing cross-infection. However, most existing deep learning-based pill recognition systems can only perform classification on classes with sufficient training data. In practice, the high cost of data annotation and the continuous increase in new pill classes necessitate the development of a few-shot class-incremental pill recognition system. This paper introduces the first few-shot class-incremental pill recognition framework, named Discriminative and Bidirectional Compatible Few-Shot Class-Incremental Learning (DBC-FSCIL). It encompasses forward-compatible and backward-compatible learning components. In forward-compatible learning, we propose an innovative virtual class synthesis strategy and a Center-Triplet (CT) loss to enhance discriminative feature learning. These virtual classes serve as placeholders in the feature space for future class updates, providing diverse semantic knowledge for model training. For backward-compatible learning, we develop a strategy to synthesize reliable pseudo-features of old classes using uncertainty quantification, facilitating Data Replay (DR) and Knowledge Distillation (KD). This approach allows for the flexible synthesis of features and effectively reduces additional storage requirements for samples and models. Additionally, we construct a new pill image dataset for FSCIL and assess various mainstream FSCIL methods, establishing new benchmarks. Our experimental results demonstrate that our framework surpasses existing State-of-the-art (SOTA) methods. The code is available at https://github.com/zhang-jinghua/DBC-FSCIL.

\end{abstract}

}

\maketitle
\IEEEdisplaynontitleabstractindextext
\IEEEpeerreviewmaketitle

\IEEEraisesectionheading{
\section{Introduction}
\label{sec:intro}}
Globally, unsafe medication practices and errors constitute a primary source of injury and preventable harm within healthcare systems, incurring an estimated annual cost of 42 billion USD~\cite{WHO}. These errors may occur at various stages of the medication process. In response to this challenge, the World Health Organization (WHO) has initiated ``Medication Without Harm'' as the theme for its third Global Patient Safety Challenge, aiming to collaborate with member states and professional institutions to mitigate these issues. Against this backdrop, the evolution of computer vision technology offers new avenues for enhancing medication safety.

APR, accurately recognizing pills by their visual appearance, is essential for ensuring patient safety and delivering effective healthcare systems, as it reduces medication dispensing errors and potential adverse drug events. APR has various potential applications. From a safety perspective, various errors can transpire at various stages of the pharmacological process. For patient care, APR can enhance treatment effectiveness in diverse situations such as disaster response, poison control interventions, and supporting patient medication adherence. For instance, poison control centers have seen increased calls for pill identification, and automated systems can alleviate the burden on experts. Additionally, consistent visual identification can aid patient persistence in medication adherence, particularly when switching between brand-name and generic drugs, as changes in a pill's appearance can influence a patient's willingness to continue therapy. In hospitals, it can strengthen the double-check process of medication dispensing by ensuring the visual identification of the medication matches the prescription exactly, effectively reducing the risk of dispensing incorrect medication to patients. Moreover, APR can also contribute to advancing remote diagnosis technologies and smart healthcare solutions.

In this context, developing an automatic prescription pill recognition system is a crucial innovation to identify oral pills from prescriptions accurately. Unlike the problem of pill recognition in application scenarios such as pill factory~\cite{mac2021application} and pill retrieval system~\cite{lee2012pill}, the prescription pill recognition system needs to simultaneously recognize pills related to unlimited classes and instances contained in one prescription. Although some studies have been directed toward developing relevant pill recognition systems, these systems are primarily based on static models trained with large datasets. However, with the introduction and evolution of pills, the classes of pills are constantly increasing, and annotating a large number of training samples to retrain recognition systems is costly. In light of these challenges, \emph{Ling et al.}~\cite{ling2020few} have crafted a pill classification framework tailored for Few-shot Learning (FSL), leveraging a combination of traditional image features and metric learning. \emph{Nguyen et al.}~\cite{nguyen2022multi} introduced an incremental multi-stream intermediate fusion framework aimed at the Class-Incremental Learning (CIL) issue. However, a blank remains in addressing the challenging but more practical few-shot class-incremental pill recognition.

To summarize, while existing efforts have marked some progress, pill recognition remains an underexploited field with the following challenges:
\begin{itemize}
\begin{figure*}[htbp!]
\centering
\includegraphics[width=0.95\textwidth]{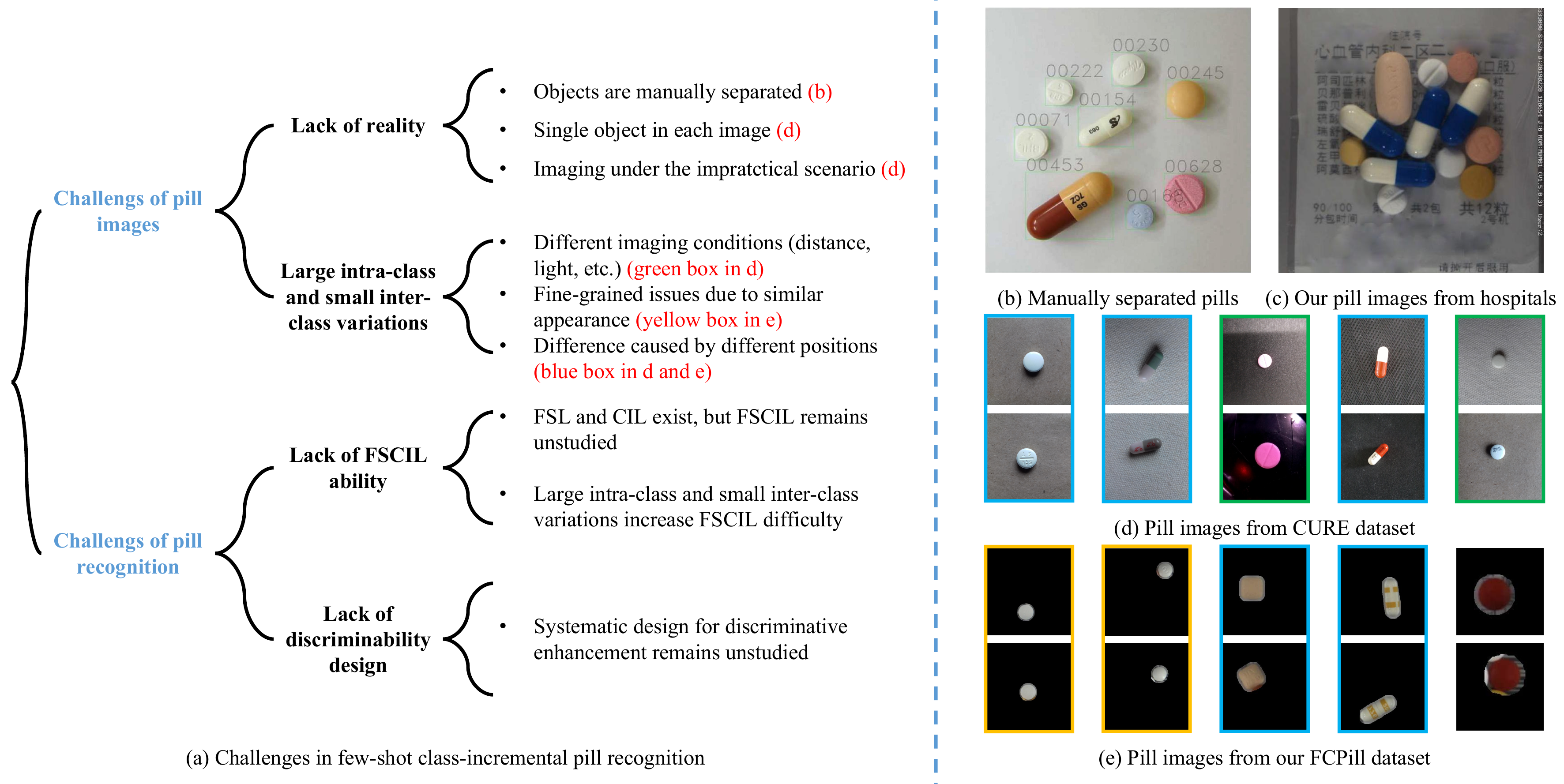}
\caption{The main challenges of few-shot class-incremental pill recognition and images from related datasets. In (a), we summarized a taxonomy of the difficulties of pill images and the design of a few-shot class-incremental pill recognition system. (b) provides the pill images used in~\cite{ou2020automatic}. These pills are manually separated, which is against the practical scenario. (c) shows the images we collected from the real application scenario in hospitals, and it can be found that the phenomenon of pill adhesion often occurs. (d) and (e) show CURE and our FCPill dataset, respectively. Each column corresponds to one class. In (d), different textures on different sides and diverse imaging conditions contribute to large intra-class variations. Conversely, (e) illustrates how pills from distinct classes can appear similar, leading to small inter-class variations. In addition, (e) highlights cases of pill adhesion.}
\label{fig1}
\end{figure*}

\item \textbf{The lack of realistic datasets.} Although some pill datasets have been proposed~\cite{chang2019deep,ou2020automatic,ling2020few}, the pills in their images are either manually separated, or these images only contain a single object, as shown in Fig.~\ref{fig1} (b) and (d). This deviates completely from the real application scenarios where pills often adhere to each other, as exemplified in Fig.~\ref{fig1}(c). The practical system is expected to be able to handle the recognition task related to randomly placed pills. The lack of realistic pill image datasets hinders the development and application of pill recognition approaches.

\item \textbf{Large intra-class variations and small inter-class variations.} Variations generated by different factors, such as randomly placed sides, illumination change, occlusion, and noise, result in significant effects on the pill appearance, causing large intra-class variations (\emph{e.g.}, examples shown in green and blue boxes of Fig.~\ref{fig1}(d) and (e)). Besides, since only a few forms are available during the pill manufacturing process (as shown in yellow boxes of Fig.~\ref{fig1}(e), many pills of different classes are white and round), small inter-class variations are common in pill images.

\item \textbf{The lack of FSCIL ability.} In pill recognition, most deep learning approaches are static, relying heavily on substantial data for predefined tasks. The emergence of new pill classes requires frequent retraining and consuming significant data, computational, and storage resources. Thus, developing a dynamic algorithm that can adapt to new pill classes with limited samples while retaining existing knowledge is crucial. Pill recognition uniquely challenges FSCIL with its large intra-class and small inter-class variations. This issue escalates as more classes are added, particularly when confusion between new and old classes is intensified by the similarity among numerous classes and the limited samples for new class learning.

\item \textbf{The lack of discriminability.} The current pill recognition research has yet to develop a systematic and comprehensive strategy to support discriminant feature learning to solve the challenge of large intra-class and small inter-class variations. This lack can affect the performance of pill recognition systems.

\end{itemize}

Our study presents the DBC-FSCIL framework tailored for few-shot class-incremental pill recognition in response to the challenges outlined. This framework encompasses two major learning strategies: forward-compatible learning during the base session and backward-compatible learning during incremental sessions. In the forward-compatible learning phase, we innovatively synthesize virtual classes and apply the CT loss function to enhance the model's ability to recognize new categories and ensure accurate recognition of previously learned categories. The strategy of synthesizing virtual classes, utilizing existing base classes to generate placeholders in the feature space for future incremental classes, enriches the diversity of the training set and provides additional semantic information to the model. This aids in better distinguishing between different feature patterns. The CT loss function further optimizes the cohesion among similar samples and the distinction between different classes, significantly boosting the model's discriminative power. For backward-compatible learning, we adopt a strategy based on uncertainty quantification to synthesize reliable pseudo-features of old classes, facilitating DR and KD. This approach allows for the flexible synthesis of any number of pseudo-features, effectively reducing the need for additional storage of samples and models and ensuring the model maintains high recognition accuracy for old categories while learning new ones. Furthermore, to support in-depth research into FSCIL, we collect and construct a new pill image dataset from real hospital environments, named FCPill. By evaluating various mainstream FSCIL methods and our method on FCPill and another public pill dataset, CURE, we establish new benchmarks for FSCIL. Our DBC-FSCIL outperforms other methods.

The key contributions of this work can be summarized as follows:

\begin{itemize}
\item We introduce a novel discriminative and bidirectional compatible FSCIL framework to perform pill recognition, DBC-FSCIL. It consists of forward compatibility learning based on virtual class generation and a novel metric loss, and backward compatibility learning based on DR and KD of pseudo-old class features. It shows superior performance across multiple benchmark datasets.

\item We propose a novel CT loss, which merges the Triplet loss and Center loss, harnessing the full potential of sample pairs for training. This approach enhances intra-class compactness and inter-class discrimination.

\item We develop an innovative KD strategy utilizing Pseudo Feature Synthesis (PFS). It uses a limited number of features from the previous session to synthesize pseudo-features. It combines uncertainty quantization and model prediction to select reliable pseudo-features for efficient KD, which exhibits better flexibility and saves storage space.

\item We construct a new pill image dataset, FCPill, for FSCIL, where the images are sampled from eight groups across seven hospitals in five cities. This dataset contains 60 base and 40 new classes, each with 600 images. In addition, a series of mainstream FSCIL methods are tested on this dataset and another public dataset, CURE, providing new benchmarks for FSCIL.

\end{itemize}

The remainders of this paper are organized as follows: In Sec.~\ref{RelatedW}, we review the relevant basic knowledge and literature; Sec.~\ref{dataset} introduce the collection and construction of our proposed datasets; Sec.~\ref{ProMethod} introduces our proposed methods in detail; Sec.~\ref{Experiments} gives dataset details and experimental results; The conclusion is provided in Sec.~\ref{Conclusion}.

\section{Related Work}
\label{RelatedW}
\subsection{Automatic Pill Recognition}
There are some existing works~\cite{lee2010pill,yu2015accurate,ou2020automatic,ling2020few,pornbunruang2022drugtionary,kwon2022deep,nguyen2022multi,chang2019deep} focusing on the development of pill recognition system. Most of these research efforts utilize private datasets, although there are instances where public datasets have been employed, notably the CURE dataset~\cite{ling2020few} and the now-unavailable NIH dataset~\cite{yaniv2016national}. In the early stage, these systems predominantly relied on feature engineering approaches~\cite{lee2010pill,yu2015accurate}, as evidenced by \cite{lee2010pill}, which highlighted the efficacy of shape, color, size, and imprint features in pill recognition. Since deep learning has shown overwhelming advantages in many computer vision tasks, it has become the dominant technology in the research of pill recognition systems. The related studies primarily focus on the object detection task and classification task. Object detection methods can be categorized into one-stage frameworks~\cite{pornbunruang2022drugtionary,kwon2022deep} and two-stage frameworks~\cite{ou2020automatic,chang2019deep}. For example, \emph{Pornbunruang et al.}~\cite{pornbunruang2022drugtionary} utilized CenterNet, a one-stage method, for direct localization and classification of pills. In contrast, \emph{Ou et al.}~\cite{ou2020automatic} implemented a two-stage approach, introducing an enhanced feature pyramid network for localization preceding classification. Recent classification studies~\cite{ling2020few,nguyen2022multi} relevant to our work have surfaced. \emph{Ling et al.}~\cite{ling2020few} developed a pill classification framework for FSL, primarily utilizing traditional image features combined with metric learning. \emph{Nguyen et al.}~\cite{nguyen2022multi} proposed an incremental multi-stream intermediate fusion framework for tackling the CIL problem. However, there remains to be a blank in addressing the challenging but more practical few-shot class-incremental pill recognition. 

\subsection{Few-shot Learning}
The purpose of FSL is to design a machine learning algorithm that can rapidly generalize to new tasks containing only a few samples with supervised information based on previous knowledge~\cite{wang2020generalizing,9733415}. Each new task usually includes a support set and a query set, and the data in the two sets do not intersect~\cite{li2022cross}. The support set contains a few labeled training samples, which assist the algorithm based on prior knowledge to make predictions on the query set. The data in the support set is often described as $N-$way $K-$shot, which means that the support set contains $N$ classes, and each class has $K$ labeled samples. In recent years, plenty of studies have focused on developing FSL algorithms. They can broadly be summarized into optimization-based and metric learning-based methods~\cite{ganea2021incremental}. Optimization-based methods train a model with data from previous base classes and then fine-tune the classifier or the whole model with data from the support set in the new task~\cite{xu2021learning}. Metric learning-based methods concentrate on learning a robust backbone to generate high-quality and transferable feature representations~\cite{zhu2022ease}. The first few-shot pill recognition framework~\cite{ling2020few} adopts the Triplet loss to obtain discriminative feature representations to perform the FSL task.

\subsection{Class-incremental Learning}
CIL aims to create algorithms that learn from new classes sequentially while retaining prior knowledge, as highlighted in~\cite{zhou2022forward,9422177}. CIL typically unfolds over sessions, each with distinct training and testing datasets. The learning shifts to the session $i$, rendering the previous session's training data inaccessible, but testing includes datasets from all sessions. FSCIL can be seen as a CIL challenge that emerges when confronted with a constrained number of samples. CIL approaches fall into three categories~\cite{zhou2022forward}: parameter-based~\cite{aljundi2018memory,zenke2017continual} focusing on preserving key model parameters, distillation-based like LwF~\cite{li2017learning} which introduced KD to CIL, and rehearsal-based~\cite{belouadah2019il2m,zhu2021prototype}, which save past data for training on new classes. iCaRL~\cite{rebuffi2017icarl} combines feature representation and classifier learning, using nearest neighbor classifiers and distillation loss to prevent forgetting.

\subsection{Few-shot Class-incremental Learning}
FSCIL is a recent machine learning topic proposed in~\cite{tao2020few}. It aims to design a machine learning algorithm that can continuously learn knowledge from a sequence of new classes with only a few labeled training samples while preserving the knowledge learned from previous classes~\cite{tao2020few,9768155}. In the FSCIL setting, the data stream usually consists of a base session with sufficient training data and a sequence of incremental sessions, whose training datasets are in the form of $N-$way $K-$shot. Previous session training datasets are unavailable once the learning process goes into session $i$. In contrast, the testing data used in session $i$ consists of the testing datasets from all previous and current sessions. Recently, quite a few works have focused on developing the FSCIL algorithm. For instance, TOPIC~\cite{tao2020few} is the first algorithm specifically used to perform the FSCIL task. It uses the neural gas network to preserve the topology of features between the base and new classes to avoid forgetting~\cite{tao2020few}. CEC~\cite{zhang2021few} adopts GAT to update the relationships between the base and new prototypes, which can help the classifier find better decision boundaries. FACT~\cite{zhou2022forward} first imported the idea of forward compatibility to enhance the adaptability of the FSCIL model for future incremental classes.

\section{FCPill Dataset}
\label{dataset}
\subsection{Data Collection}
Existing pill datasets are categorized into single-object and multi-object types, with the latter often manually separating pills, which does not align with real-world applications. In most hospitals, patients usually receive pill packets based on prescriptions, typically containing multiple classes and instances of pills that are often adhered or overlapped. This scenario is not accurately represented in current datasets. To fill this gap, we collected a substantial number of original pill packet images and their corresponding electronic prescriptions from real hospitals. Following approval from the relevant ethics committee, we employed a device equipped with top and bottom cameras to capture images of the pill packets, recording both the front and back views. Ultimately, we gathered 473,148 original pill packet images from eight groups across seven hospitals in five Chinese cities. Images containing the same pill class were categorized based on the prescriptions to facilitate dataset construction.

\subsection{Data Annotation}
After acquiring images of each pill class, we started to extract individual pill objects. Given that most pill packet images contain multiple pills, we first located and segmented individual pills, followed by manual classification based on pill templates. Considering the real scenarios, where pills often adhere to each other, we primarily employed watershed algorithms and dilation-erosion processes for segmentation. To preserve complete pill information as much as possible, we dilated the segmentation results to cover the entire object area, thereby maintaining the challenge posed by adhesion. After segmentation, trained human experts performed data annotation, comparing pill templates to select the correct individual pill images. After completing the preliminary annotation, we followed the commonly used FSCIL benchmark dataset, miniImageNet, to filter out classes with a minimum of 600 samples. This criterion led to the selection of 100 classes, forming the foundation of our FCPill dataset.

\subsection{Evaluation Protocol}
After compiling the FCPill dataset comprising 100 classes, each with 600 samples, we developed an evaluation protocol tailored for FSCIL. Specifically, similar to miniImageNet, we partitioned the dataset, allocating 300 images per class for training and the remaining 300 for testing. After that, 60 classes were earmarked as base classes, while the rest were categorized as incremental classes. These incremental classes were further divided into 8 incremental sessions, each incorporating 5 classes. Within each class of these sessions, 5 samples were randomly chosen to serve as the training data, adhering to the 5-way 5-shot format. The specific partition file can refer to our code.

\section{Proposed Methodology}
\label{ProMethod}
In this section, the problem setting of FSCIL is first introduced. Subsequently, an overview of our proposed framework is presented. This is followed by detailed descriptions of the forward-compatible learning and the backward-compatible learning within the framework.

\subsection{Problem Setting}
Assume $\{D^0_{train},\cdots,D^n_{train}\}$ and $\{D^0_{test},\cdots,D^n_{test}\}$ denote the training and testing datasets in FSCIL sessions. The $n$ means the incremental session number in the current FSCIL task. $D^0_{train}$ denotes the training dataset in the base session, which contains abundant labeled training data. $\forall$ integer $i\in[1,n], D^i_{train}$ is in form of $N-$way $K-$shot, which means the training dataset in session $i$ contains $N$ classes and each class has $K$ labeled samples. $D^i_{test}$ denotes the testing dataset in session $i$. $\forall$ integer $i\in[0,n]$, $C^i$ denotes the corresponding label space of $D^i_{train}$ and $D^i_{test}$. The relevant classes in different sessions have no intersection, \emph{i.e.}, $\forall$ integer $i,j\in[0,n]$ and $i\not=j$, $C^i {\cap} C^j = \emptyset$. When the training process comes into session $i$, only the entire $D^i_{train}$ is available, while the entire training datasets of previous sessions are no longer available. For the evaluation at session $i$, the testing data consists of all the testing datasets from current and previous sessions, \emph{i.e.}, $D^0_{test} \cup \cdots \cup D^i_{test}$. FSCIL encounters two major challenges: unreliable empirical risk minimization due to limited supervised data, affecting model generalization and increasing overfitting risks, and the stability-plasticity dilemma, where continual addition of new classes risks overwriting old knowledge, leading to catastrophic forgetting or intransigence. Balancing model stability and plasticity remains a central challenge~\cite{zhang2023few}.

\subsection{Overall Framework}

\begin{figure}[htbp!]
\centering
\includegraphics[width=0.49\textwidth]{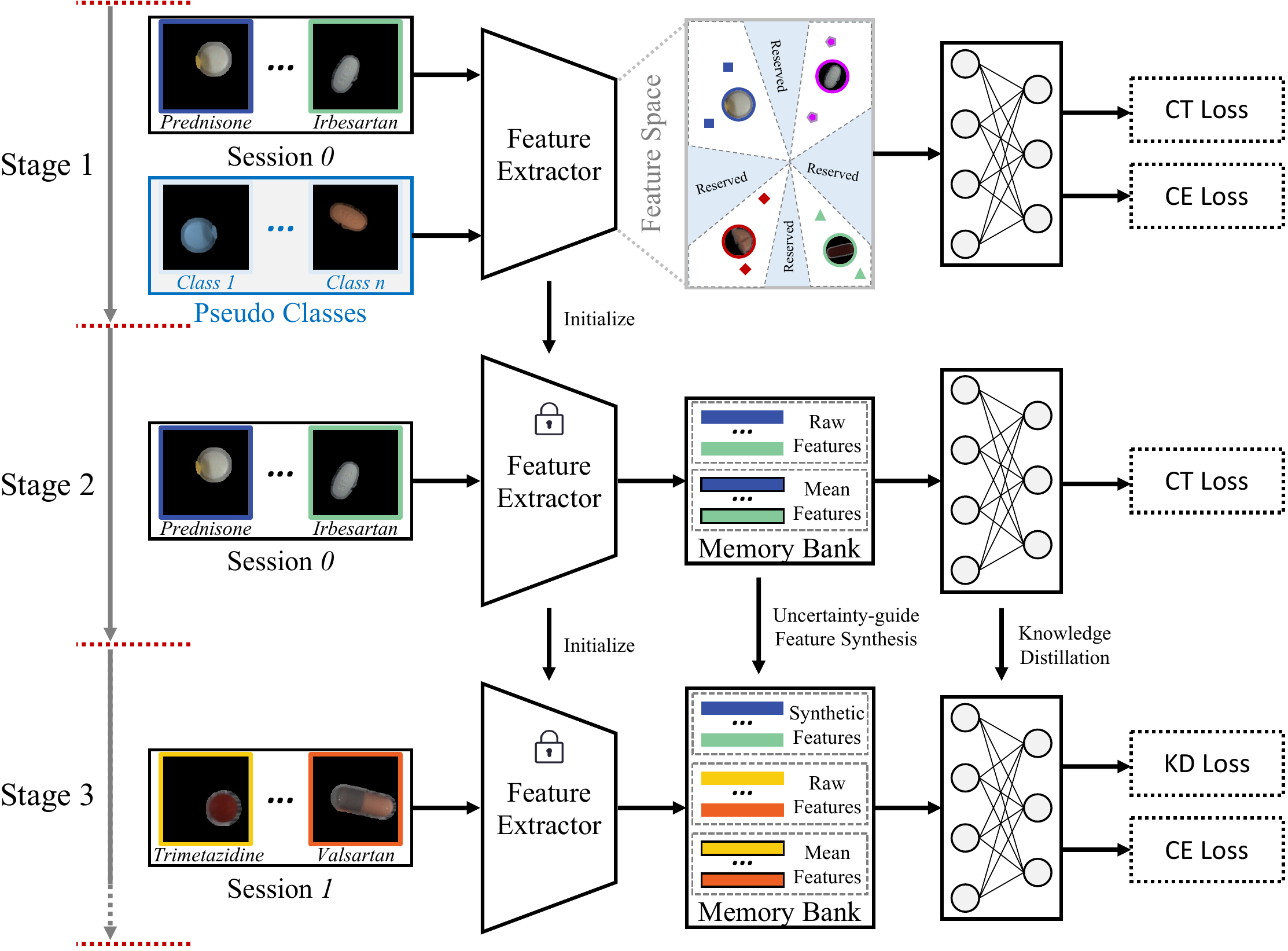}
\caption{The DBC-FSCIL Framework for Pill Recognition: Stage 1 focuses on the virtual class generation and forward-compatible learning; Stage 2 aims at fine-tuning the model for base session classification; Stage 3 is dedicated to the uncertainty-guided synthesis of pseudo old class features for KD, ensuring backward compatibility in the incremental learning process.}
\label{Fig2}
\end{figure}

The concept of compatibility is a design characteristic considered in software engineering~\cite{zhou2022forward}. Forward compatibility enables a system to process inputs intended for future versions, whereas backward compatibility facilitates interoperability with older legacy systems~\cite{zhou2022forward}. To perform the few-shot class-incremental pill recognition with the continuous learning of new classes and the retention of old class knowledge, we propose a novel bidirectional compatible FSCIL framework (DBC-FSCIL), as depicted in Fig.\ref{Fig2}. DBC-FSCIL comprises two key components: forward-compatible learning, covered in phases 1 and 2, and backward-compatible learning in phase 3, as outlined in Fig.\ref{Fig2}. 

In short, in Stage 1, the DBC-FSCIL framework generates virtual classes using existing pill images. These virtual classes are combined with the real base classes for training the backbone, endowing its forward-compatible capability. This enables efficient feature extraction for future incremental classes, allowing the model to adapt to new classification tasks with limited samples while low interference with previous classification tasks. Considering the large intra-class and small inter-class variations in pill images, we propose a novel metric loss function to facilitate learning more discriminative features by the backbone. Stage 2 focuses on further fine-tuning the real classes in the base session after the weight initialization, enhancing its adaptability for the classification in the base session. Stage 3 aims to endow the model with backward compatibility during the incremental learning phase. This is achieved by employing uncertainty quantification and model predictions to synthesize and select reliable pseudo-features of old classes. These pseudo-features are then integrated into the incremental learning process through DR and KD strategies, thereby effectively preserving previous knowledge.

\subsection{Forward Compatibility Learning}
Many FSCIL algorithms adopt a strategy of freezing the backbone after base session training to mitigate overfitting and catastrophic forgetting in the incremental learning process. They then integrate the frozen backbone with the Nearest Class Mean (NCM) classifier to perform FSCIL. This strategy surpasses many other methods~\cite{shi2021overcoming}. Specifically, in the base training phase, Cross-entropy (CE) loss is prevalently utilized to initialize the backbone. The total classification loss is expressed as:
\begin{equation}
\label{eq1}
\mathcal{L}_{cls}\left( \phi;\bm{{\rm x}},y \right) = \mathcal{L}_{ce}\left( {\phi\left( \bm{{\rm x}} \right),y} \right),
\end{equation}
where $\mathcal{L}_{ce}\left( {\cdot} \right)$ denotes the CE loss, $\bm{{\rm x}}$ represents the sample, $y$ is the corresponding label, and the model can be decomposed into the backbone and classifier: $\phi(\bm{{\rm x}}) = W^{T}g(\bm{{\rm x}})$, with $\phi(\bm{{\rm x}}) \in \mathbb{R}^{\left| \mathcal{C}^{0} \right| \times 1}$, $g(\bm{{\rm x}}) \in \mathbb{R}^{d \times 1}$, and $W \in \mathbb{R}^{d \times \left| \mathcal{C}^{0} \right|}$. After the base training, the backbone is frozen. The prototypes in the NCM classifier are generated by averaging the features of each class, formulated as $W = \{ {\bm{w}_{1}^{0},\bm{w}_{2}^{0},\cdots,\bm{w}_{|\mathcal{C}^{0}|}^{0}} \} \cup \cdots \cup \{ {\bm{w}_{1}^{i},\cdots,\bm{w}_{|\mathcal{C}^{i}|}^{i}} \}$. During inference, the features derived from the test sample are utilized to compute similarity with each prototype in the NCM classifier, typically employing cosine similarity. The classification is then based on the prototype most similar to the test sample. 

In the base session, despite the availability of sufficient training samples for initializing the backbone, the robustness and generalizability of the backbone, primarily trained by the CE loss for classification, remain ambiguous for unseen incremental classes~\cite{elsayed2018large}. Moreover, unlike other natural images, pill images exhibit the challenge of large intra-class and small inter-class variations, exacerbating the issue of poor generalization. To address this challenge, we propose a strategy that involves optimizing the loss function to promote the learning of discriminative features and synthesizing virtual categories to enhance the forward compatibility of the model.

\subsubsection{Center-Triplet Loss}
The features learned by CE loss often lack discriminability, primarily because the core objective of CE loss is to delineate decision boundaries between different classes~\cite{he2018triplet}. Studies~\cite{elsayed2018large,liu2016large} have shown that due to the poor class margins, it exhibits poor generalization performance, potentially rendering it unsuitable for FSCIL. Considering that the FSCIL model needs to continuously adapt to new classes, if the prototypes are sufficiently distant from each other and the features of the same class are closely clustered around their prototype, it facilitates better support for FSCIL. In many FSL studies~\cite{ling2020few,zhou2020siamese,shi2021conditional}, triplet loss is widely used to learn discriminative feature representations. However, it cannot consider the inter-distance within one class~\cite{he2018triplet}, and the triplet mining significantly affects the training process. 

Specifically, for $\forall \left(\bm{{\rm x}}_{a},\bm{{\rm x}}_{p},\bm{{\rm x}}_{n}\right)\in \mathcal{T}$, the optimization purpose of triplet loss is 
\begin{equation}
\label{eq2}
\Vert{g\left(\bm{{\rm x}}_{a}\right)-g\left(\bm{{\rm x}}_{p}\right)}\Vert+m<\Vert{g\left(\bm{{\rm x}}_{a}\right)-g\left(\bm{{\rm x}}_{n}\right)}\Vert,
\end{equation}
where $\mathcal{T}$ denotes the set of all the possible triplets of training samples. Each triplet consists of an anchor sample $\bm{{\rm x}}_{a}$, a positive sample $\bm{{\rm x}}_{p}$ with the same label as $\bm{{\rm x}}_{a}$, and a negative sample $\bm{{\rm x}}_{n}$ from another class. $m$ is the margin value. $\Vert\cdot\Vert$ denotes the calculation of Euclidean distance. Based on the optimization purpose in Eq.~\ref{eq3}, the Triplet loss can be formulated as follows:
\begin{equation}
\label{eq3}
\begin{aligned}
\mathcal{L}_{t}\left( g;\bm{{\rm x}} \right)= \max(0, m+ \Vert{g\left(\bm{{\rm x}}_{a}\right)-g\left(\bm{{\rm x}}_{p}\right)}\Vert-\Vert{g\left(\bm{{\rm x}}_{a}\right)-g\left(\bm{{\rm x}}_{n}\right)}\Vert).
\end{aligned}
\end{equation}
It's worth noting that not all the triplets in $\mathcal{T}$ can promote model optimization. Most triplets in $\mathcal{T}$ belong to the easy triplets that satisfy Eq.~\ref{eq3} and cannot contribute to the training process. Only semi-hard and hard triplets in $\mathcal{T}$ facilitate model progress. However, extremely hard triplets can cause model collapse~\cite{schroff2015facenet}. The effectiveness of triplet mining is crucial, but some valuable triplets are often overlooked~\cite{he2018triplet}. For example, a triplet $A$ with positive pair distance $b$ and negative pair distance $a$ is considered easy and ineffective if the margin in Triplet loss, $m$, is less than $\left|b-a\right|$. Similarly, a triplet $B$ with distances $b+n$ and $a+n$ (where $n$ is a positive number) remains ineffective. This scenario shows that Triplet loss inadequately constrains intra-class compactness, leading to the neglect of triplets that could be beneficial, suggesting the need for better utilization of potential triplets to improve intra-class compactness.

\begin{figure}[htbp!]
\centering
\includegraphics[width=0.43\textwidth]{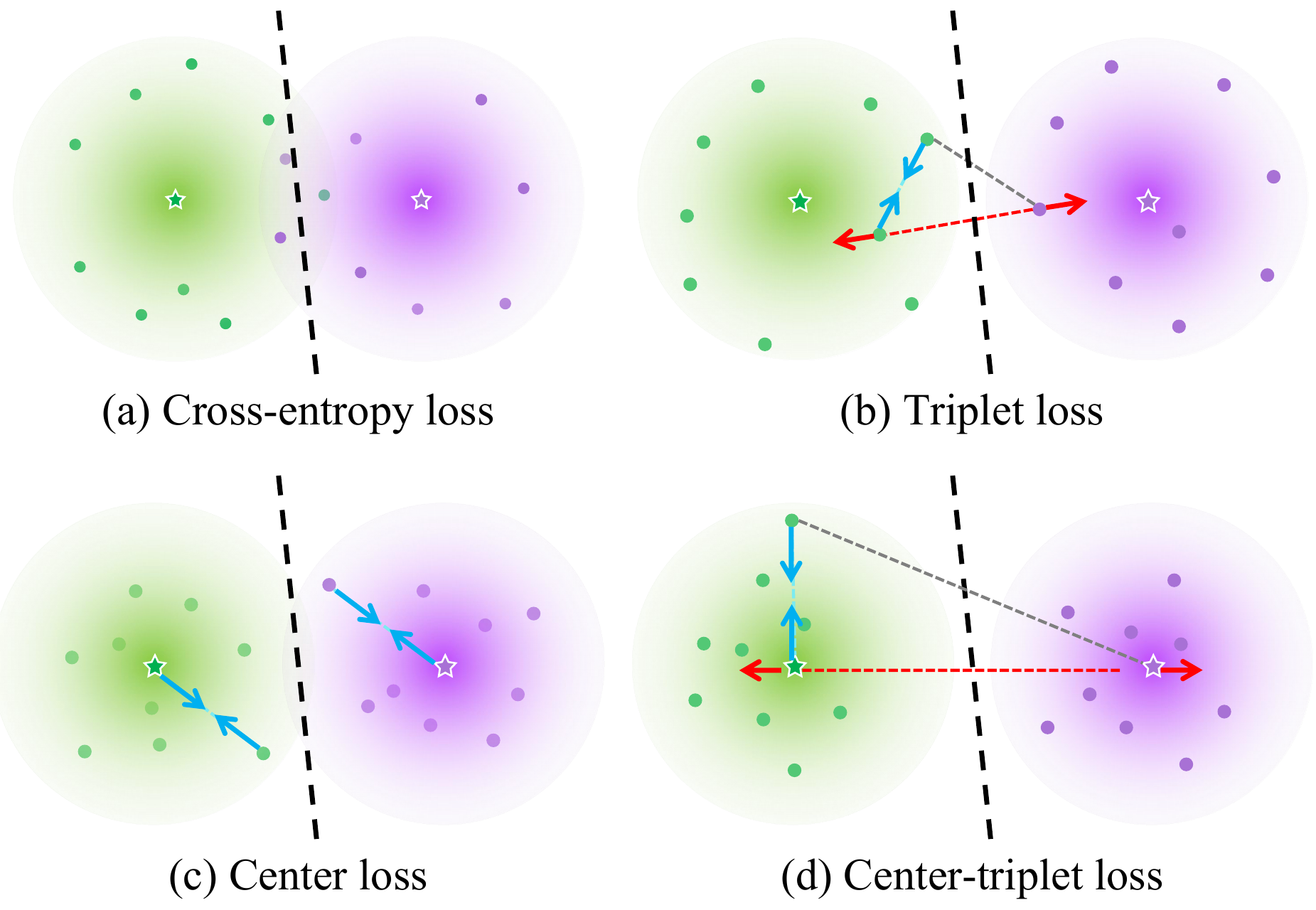}
\caption{The illustrations of related metric losses. (a) CE loss aims to learn the decision boundary; (b) Triplet loss seeks to constrain the distance between features of the same class to be less than the distance between different classes by a predefined margin; (c) Center loss encourages the intra-class compactness; (d) Our proposed CT loss further promote the intra-class compactness and inter-class separability by directly considering the distance between different class centers.}
\label{Fig3}
\end{figure}

To overcome Triplet loss's limitations, we propose the CT loss, combining Center loss concepts to boost intra-class compactness. It aims at distinct representation learning by anchoring a class and minimizing the distance between its samples and center, less than the distance to the nearest different class center. The CT loss is mathematically expressed as:
\begin{equation}
\label{eq4}
\begin{aligned}
\mathcal{L}_{ct}\left( g;\bm{{\rm x}} \right)= \max(0, m+ \Vert g\left(\bm{{\rm x}}\right)-\bm{{\rm c}}_{y}\Vert-\min_{j\neq y}\Vert \bm{{\rm c}}_{y}-\bm{{\rm c}}_{j}\Vert
).
\end{aligned}
\end{equation}
Here, the triplet $\left(\bm{{\rm x}},\bm{{\rm c}}_{y},\bm{{\rm c}}_j\right)$ includes a sample $\bm{{\rm x}}$, its class center $\bm{{\rm c}}_{y}$, and the nearest different class center $\bm{{\rm c}}_j$. The loss updates the backbone to ensure the distance between the sample's feature and its center $\bm{{\rm c}}_{y}$ is less than the distance to $\bm{{\rm c}}_{y}$ by a margin $m$. Fig.~\ref{Fig3} shows the details of related losses. The joint training uses both CE loss and CT loss, formulated as:
\begin{equation}
\label{eq5}
\begin{aligned}
\mathcal{L}_{total}=\mathcal{L}_{cls}+\lambda \mathcal{L}_{ct},
\end{aligned}
\end{equation}
where $\lambda$ balances the CT and classification losses.

\subsubsection{Virtual Class Generation}
In addition to the development of CT loss for enhanced discriminative feature learning, our approach during the base training is focused on establishing forward compatibility. Similar to existing works~\cite{zhou2022forward,peng2022few,song2023learning}, during the base training, we synthesize various virtual classes and combine them with base classes to train the backbone, ensuring its generalization capability in incremental sessions. 
These virtual classes play two crucial roles in achieving forward compatibility: they act as placeholders in the feature space for future class updates and serve as sources of diverse semantic knowledge, guiding the model to engage in extensive learning across various semantics. 

Specifically, due to the significant differences between pill images and natural images, the methods of synthesizing virtual classes in existing forward compatibility studies are not able to accurately emulate real pill classes. By analyzing the characteristics of pill images, we observed that pills have structured shapes, such as circular, oval, or capsule forms, and exhibit uniform color, fixed sizes, and consistent textures within each class. Leveraging these characteristics, we develop a specialized approach to synthesize virtual pill classes. This involves applying random color and size transformations to pill samples in the base session, thereby generating realistic virtual classes. The generation process can be expressed as $\left( {\bm{{\rm x}}_{rv},~y_{rv}} \right) = \mathcal{F}\left( {\bm{{\rm x}},~y} \right)$, where $\left( {\bm{{\rm x}}_{rv},~y_{rv}} \right)$ denotes the image-label pair from the union of real and virtual dataset, and $\mathcal{F}$ denotes the transformation function. This strategy doubles the label space and ensures the virtual classes closely resemble actual pill classes. The union of real and virtual classes is used to train the backbone to facilitate forward compatibility and generalization capability. Following the integration of virtual classes, the training loss functions for the backbone can be articulated as follows:
\begin{equation}
\label{eq6}
\begin{split}
&\mathcal{L}_{cls}\left( {\phi;\bm{{\rm x}}_{rv},y_{rv},\mathcal{F}} \right) = \mathcal{L}_{ce}\left( {\phi\left( \bm{{\rm x}}_{rv} \right),y_{rv}} \right),\\
&\mathcal{L}_{ct}\left( f;\bm{{\rm x}}_{rv} \right)= \max(0, m+ \Vert f\left(\bm{{\rm x}}_{rv}\right)-\bm{{\rm c}}_{y_{rv}}\Vert-\min_{j\neq y_{rv}}\Vert \bm{{\rm c}}_{y_{rv}}-\bm{{\rm c}}_{j}\Vert),\\
&\mathcal{L}_{total}=\mathcal{L}_{cls}+\lambda \mathcal{L}_{ct}.
\end{split}
\end{equation}

\subsection{Backward Compatibility Learning}
In our DBC-FSCIL framework, backward-compatible learning primarily aims to efficiently retain previous knowledge while learning new knowledge, achieved through DR and KD. Currently, although some FSCIL methods~\cite{kukleva2021generalized,dong2021few} balance the learning of new and old knowledge by employing raw DR and KD during the learning process of new sessions, they typically require storing samples from previous classes and face the challenge of insufficient comprehensive retention of old knowledge due to limited available samples. Specifically, existing methods encounter several challenges: 1) Limited samples in incremental sessions may hinder a comprehensive review of knowledge; 2) As the number of classes increases, so does the demand for storage space; 3) Storing raw samples introduces additional privacy risks; 4) KD necessitates extra storage for models trained in previous sessions. Our novel strategy addresses these challenges by merging DR and KD with PFS, enhancing backward compatibility, and reducing storage demands.

\subsubsection{Pseudo Feature Synthesis}
Unlike existing methods, we train a robust backbone in the base session and keep its feature extractor frozen during incremental sessions, training only the fully connected layers to maintain generalization capabilities and forward compatibility. We consider using features extracted by this frozen feature extractor for DR and KD, aiming to reduce storage space requirements and minimize privacy risks. However, due to the few-shot condition, relying solely on these features might not be sufficient for a comprehensive knowledge review. Therefore, we propose a strategy that utilizes features extracted from existing classes combined with uncertainty quantification and model predictions to synthesize and select reliable pseudo-features, thus overcoming the limitations imposed by the few-shot samples. The details are showcased in Alg.~\ref{alg1}.

\begin{algorithm}[htb]
\begin{spacing}{1}
\caption{Pseudo-feature Synthesis for Session $i$}
\label{alg1}
\SetAlgoLined
\KwIn{Training dataset $D^i_{train}$, trained model $\phi_i$ (with feature extractor $\varphi_i$ and fully connected layers $\psi_i$), number of features $P$ to be stored per class, number of pseudo-features $Q$ to synthesize per class.}
\KwOut{Set of synthesized pseudo-features $S$.}
Initialize an empty set $S$ for storing pseudo-features.\\
\ForEach{class $c \in D^i_{train}$}{
    Extract features $\{f_{c_i}\}_{i=1}^{N_c}$ for all $N_c$ samples in class $c$ using $\varphi_i$.\\
    Compute mean feature $\mu_c = \frac{1}{N_c}\sum_{i=1}^{N_c} f_{c_i}$.\\
    Store $P$ randomly selected features and mean feature $\mu_c$ in a memory bank $M_c$.
}
\ForEach{class $c \in D^i_{train}$}{
    Set $count_c = 0$.\\
    \While{$count_c < Q$}{
        Select a random feature vector $f$ from $M_c$.\\
        Generate a random scalar $\alpha \in (0, 1)$.\\
        Synthesize pseudo-feature $f_v = \alpha f + (1 - \alpha) \mu_c$.\\
        Predict class label of $f_v$ using $\psi_i(f_v)$ and compute information entropy $H(f_v)$.\\
        \If{$\psi_i(f_v)$ predicts class $c$ and $H(f_v) < \text{threshold}$}{
            Append $f_v$ to $S$.\\
            Increment $count_c$ by 1.
        }
    }
}
\end{spacing}
\end{algorithm}

In Alg.~\ref{alg1}, information entropy, also called Shannon entropy, is a well-defined measurement for uncertainty, which can be defined as follows:
\begin{equation}
\label{eq7}
H(f_v) = -\sum_{c=1}^{|\mathcal{C}|} \psi_i(f_v)\log \psi_i(f_v),
\end{equation}
where $|\mathcal{C}|$ denotes the total number of classes in session $i$, $\psi_i(f_v)$ represents the probability. High entropy denotes low confidence and vice versa. Our PFS approach allows the synthesis of a specified number $Q$ of reliable pseudo-features for each class. This strategy conserves sample storage space, mitigates privacy concerns, and facilitates the simulation of more diverse features. Consequently, this ensures the comprehensive review of previous knowledge during DR and enhances the preservation of previous knowledge in the KD process.

\subsubsection{Data Reply and Knowledge Distillation}
Following the synthesis of reliable pseudo-features from previous sessions, our DBC-FSCIL framework employs two strategies to ensure a balanced integration of old and new knowledge. Initially, we utilize the pseudo-features representing old categories for DR. This approach ensures that while the fully connected layers of the model are learning features of new classes in session $t$, they are also engaged in learning from the pseudo-features of old classes, thereby maintaining a balance in knowledge acquisition across old and new classes. This process is mathematically represented as:
\begin{equation}
\label{eq8}
\mathcal{L}_{cls}\left(\psi_t;f_{rv},y_{rv} \right) = \mathcal{L}_{ce}\left( {\psi_t\left(f_{rv}\right),y_{rv}} \right),
\end{equation}
where $\psi_t$ represents the fully connected layers, $f_rv$ denotes the union of features extracted from the current session and pseudo-features of previous sessions, and $y_{rv}$ is the corresponding label. In addition to the DR, we have incorporated a KD approach in our framework. This technique facilitates the transfer of knowledge learned from old to new models. The distillation process is mathematically represented using the Kullback-Leibler (KL) divergence, a measure of how one probability distribution diverges from a second, expected probability distribution:
\begin{equation}
\label{eq9}
\mathcal{L}_{distill}\left(\psi_t,\psi_{t-1};f_{rv} \right) = KL\left( \frac{\psi\left(f_{rv}\right)}{T},  \frac{\psi_{t-1}\left(f_{rv}\right)}{T} \right)
\end{equation}
where $T$ is the temperature parameter. The KL divergence in this context quantifies the difference between the softened probability distributions of the teacher and student models. The joint training loss in incremental sessions can be formulated as follows:
\begin{equation}
\label{eq10}
\begin{aligned}
\mathcal{L}_{total}= \mathcal{L}_{cls}+\beta \mathcal{L}_{distill},
\end{aligned}
\end{equation}
where $\beta$ balances the classification and KD losses.

The flexibility in the number of synthesized pseudo-features further enhances the efficiency of both DR and KD. Moreover, as our framework utilizes a consistent feature extractor across different sessions, it necessitates only the retention of weights in the fully connected layers to assist in the KD process. This significantly reduces the memory space required for model storage. Our method offers an efficient solution for the effective amalgamation of old and new knowledge in FSCIL scenarios.

\section{Experiments}
\label{Experiments}
\subsection{Datasets and Protocols}
Our DBC-FSCIL framework is evaluated on our proposed FCPill dataset and the public CURE pill dataset~\cite{ling2020few}. 

\textbf{\emph{m}CURE:}
In addition to our proposed FCPill dataset, we evaluate our framework on CURE~\cite{ling2020few}, a public pill image dataset originally proposed for FSL. CURE contains 1873 images of 196 classes, and each class has approximately 45 samples. To make it fit the setting of FSCIL, we follow the similar splits as \emph{mini}ImageNet in~\cite{tao2020few} to sample 171 classes to create the \emph{m}CURE dataset, where 171 classes are divided into 91 base classes and 80 new classes. These new classes are further divided into eight incremental sessions, and the training data in each session is in the form of 10-way 5-shot.

To comprehensively evaluate our DBC-FSCIL framework, we employed three metrics: 1) The accuracy values obtained on every session; (2) The Performance Drop (PD) rate, which measures the absolute decline in accuracy from the base to the final session; (3) The Average Accuracy (AA) of all sessions.

\subsection{Implementation Details}
\textbf{Model Configurations:}
In most FSCIL studies, ResNet18 is frequently utilized as the backbone. Our research also employs ResNet18 as the backbone for the pill datasets. A distinctive aspect of our approach is the addition of a fully connected layer at the end of ResNet18 to enhance backward-compatible training during the incremental learning stage. During the training stage, the model is optimized using SGD with a learning rate of 0.1, momentum of 0.9, and weight decay of 0.0005. After completing the base session training, we freeze all parameters except those in fully connected layers, which are exclusively trained during incremental sessions. Classification is performed using the softmax function. Our framework is implemented in PyTorch 2.1 and Python 3.9, and trained on the Nvidia Tesla V100 GPU.

\textbf{Training Details:}
In the experiment for pill datasets, the training framework is divided into three stages: forward-compatible training in the base session, fine-tuning for classification in the base session, and backward-compatible training in incremental sessions. For forward-compatible training, we set the epoch number to 100, CT loss weights to 0.05 (for FCPill) and 0.1 (for \emph{m}CURE), and margin values to 1 (for FCPill) and 2 (for \emph{m}CURE). The fine-tuning phase involves 50 epochs with 5 (for FCPill) and 4 (for \emph{m}CURE) random stored features per class. In the backward-compatible training phase, we set the epoch number to 50, synthesize 10 (for FCPill) and 12 (for \emph{m}CURE) pseudo-features per class, with KD loss weights of 0.4 (for FCPill) and 0.6 (for \emph{m}CURE) and a distillation temperature of 3 (for FCPill) and 5 (for \emph{m}CURE).

\subsection{Comparison with Other Methods}

To fully demonstrate the performance of our method, we conduct performance comparisons on our proposed FCPill and the public \emph{m}CURE with several representative SOTA methods. These include compatibility methods such as FACT~\cite{zhou2022forward}, ALICE~\cite{peng2022few}, and SAVC~\cite{song2023learning}, as well as non-compatibility methods like CEC~\cite{zhang2021few}, LIMIT~\cite{zhou2022few}, SSFE-Net~\cite{pan2023ssfe}, and BiDistFSCIL~\cite{zhao2023few}.

\begin{figure}[htbp!]
\centering
\includegraphics[width=0.495\textwidth]{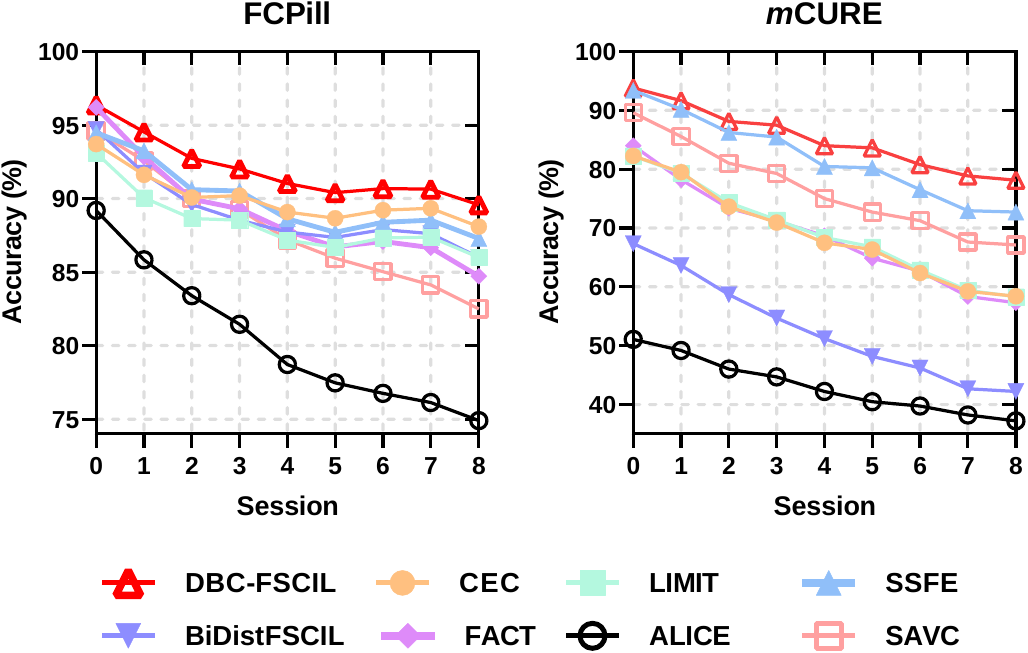}
\caption{Comparison with SOTA methods on FCPill and \emph{m}CURE. Our method, DBC-FSCIL, comprehensively surpasses other methods.}
\label{Fig4}
\end{figure}

\textbf{Results on FCPill:}
Fig.~\ref{Fig4} and Tab.~\ref{tab1} show that our DBC-FSCIL framework consistently delivers the best performance in all sessions on the FCPill dataset. In the base session, it achieves a $96.38\%$ accuracy, outperforming the leading non-compatibility method, BiDistFSCIL, by $1.67\%$. Against other compatibility methods, our framework surpasses FACT by $0.16\%$, ALICE by $7.18\%$, and SAVC by $1.76\%$. This success indicates that our forward-compatible strategy effectively enhances class discriminability through virtual class generation and CT loss. Compared to existing compatibility methods, our framework not only adopts forward compatibility but also focuses on sustaining backward compatibility during incremental learning. In the final session, our method achieves the best accuracy of $89.59\%$, and it exceeds forward-compatible methods, including FACT by $4.86\%$, ALICE by $14.68\%$, and SAVC by $7.09\%$. Moreover, it surpasses non-compatibility methods in final session performance. Regarding AA, our framework substantially surpasses the best-performing compatibility and non-compatibility methods by $2.97\%$ and $2.00\%$, respectively. Although our PD rate stands at $6.79\%$, which is the second-best performance, it shows the strongest resistance to catastrophic forgetting among compatibility methods. It's important to note that, as some studies~\cite{peng2022few} suggest, PD is not an exhaustive metric for evaluating resistance to forgetting.

\begin{table*}[htbp!]
\centering
\renewcommand{\arraystretch}{1.2}
\setlength\tabcolsep{4.8pt}
\caption{Comparison results of our DBC-FSCIL against other SOTA methods on FCPill and \emph{m}CURE. Comp. shows whether the method is compatible. (In \%).} 
\begin{tabular}{ccccccccccccccc}
\hline
\multirow{2}{*}{Dataset}                                             & \multirow{2}{*}{Comp.} & \multirow{2}{*}{Method} & \multirow{2}{*}{Venue} & \multicolumn{9}{c}{Session ID}                                                                                                                         & \multirow{2}{*}{AA$\uparrow$} & \multirow{2}{*}{PD$\downarrow$} \\ \cline{5-13}
                                                                     &                        &                         &                        & 0              & 1              & 2              & 3              & 4              & 5              & 6              & 7              & 8              &                     &                     \\ \hline
\multirow{8}{*}{FCPill}                                              & \multirow{4}{*}{No}    & CEC~\cite{zhang2021few}                     & \small\textit{CVPR 21}                & 93.71          & 91.63          & 90.08          & 90.22          & 89.10          & 88.67          & 89.22          & 89.34          & 88.11          & 90.01               & \textbf{5.59}       \\
                                                                     &                        & LIMIT~\cite{zhou2022few}                   & \small\textit{TPAMI 22}               & 93.11          & 90.07          & 88.65          & 88.54          & 87.18          & 86.68          & 87.33          & 87.39          & 86.01          & 88.33               & 7.10                \\
                                                                     &                        & SSFE-Net~\cite{pan2023ssfe}                    & \small\textit{WACV 23}                & 94.49          & 93.26          & 90.61          & 90.53          & 88.63          & 87.72          & 88.40          & 88.54          & 87.29          & 89.94               & 7.20                \\
                                                                     &                        & BiDistFSCIL\cite{zhao2023few}             & \small\textit{CVPR 23}                & 94.71          & 91.74          & 89.61          & 88.54          & 87.73          & 87.36          & 87.90          & 87.62          & 86.00          & 89.02               & 8.71                \\ \cline{2-15} 
                                                                     & \multirow{4}{*}{Yes}   & FACT~\cite{zhou2022forward}                   & \small\textit{CVPR 22}                & 96.22          & 92.84          & 89.98          & 89.31          & 87.80          & 86.72          & 87.09          & 86.67          & 84.73          & 89.04               & 11.49               \\
                                                                     &                        & ALICE~\cite{peng2022few}                     & \small\textit{ECCV 22}                & 89.20          & 85.84          & 83.40          & 81.46          & 78.73          & 77.48          & 76.76          & 76.13          & 74.91          & 80.43               & 14.29               \\
                                                                     &                        & SAVC~\cite{song2023learning}                    & \small\textit{CVPR 23}                & 94.62          & 92.57          & 90.02          & 89.28          & 87.20          & 85.95          & 85.04          & 84.14          & 82.50          & 87.92               & 12.12               \\ \cline{3-15} 
                                                                     &                        & \textbf{DBC-FSCIL}      & -                      & \textbf{96.38} & \textbf{94.54} & \textbf{92.74} & \textbf{92.03} & \textbf{91.04} & \textbf{90.41} & \textbf{90.68} & \textbf{90.66} & \textbf{89.59} & \textbf{92.01}      & 6.79                \\ \hline
\multirow{8}{*}{\emph{m}CURE} & \multirow{4}{*}{No}    & CEC~\cite{zhang2021few}                     & \small\textit{CVPR 21}                & 82.26          & 79.52          & 73.65          & 70.92          & 67.52          & 66.35          & 62.36          & 59.24          & 58.40          & 68.91               & 23.86               \\
                                                                     &                        & LIMIT~\cite{zhou2022few}                   & \small\textit{TPAMI 22}               & 82.26          & 79.33          & 74.35          & 71.40          & 68.44          & 66.79          & 62.87          & 59.42          & 58.31          & 69.24               & 23.95               \\
                                                                     &                        & SSFE-Net~\cite{pan2023ssfe}                    & \small\textit{WACV 23}                & 93.41          & 90.22          & 86.24          & 85.46          & 80.48          & 80.27          & 76.53          & 72.95          & 72.71          & 82.03               & 20.70               \\
                                                                     &                        & BiDistFSCIL\cite{zhao2023few}             & \small\textit{CVPR 23}                & 67.36          & 63.66          & 58.69          & 54.71          & 51.22          & 48.19          & 46.23          & 42.70          & 42.22          & 52.78               & 25.14               \\ \cline{2-15} 
                                                                     & \multirow{4}{*}{Yes}   & FACT~\cite{zhou2022forward}                    & \small\textit{CVPR 22}                & 84.00          & 78.24          & 73.39          & 71.20          & 68.67          & 64.90          & 62.68          & 58.39          & 57.31          & 68.75               & 26.69               \\
                                                                     &                        & ALICE~\cite{peng2022few}                   & \small\textit{ECCV 22}                & 51.10          & 49.21          & 46.04          & 44.71          & 42.21          & 40.46          & 39.74          & 38.23          & 37.22          & 43.21               & \textbf{13.88}      \\
                                                                     &                        & SAVC~\cite{song2023learning}                    & \small\textit{CVPR 23}                & 89.63          & 85.57          & 81.04          & 79.28          & 75.04          & 72.72          & 71.26          & 67.64          & 67.14          & 76.59               & 22.49               \\ \cline{3-15} 
                                                                     &                        & \textbf{DBC-FSCIL}      & -                      & \textbf{93.85} & \textbf{91.67} & \textbf{88.15} & \textbf{87.48} & \textbf{84.00} & \textbf{83.63} & \textbf{80.81} & \textbf{78.86} & \textbf{78.15} & \textbf{85.18}      & 15.69               \\ \hline
\end{tabular}
\label{tab1}
\end{table*}

\textbf{Results on \emph{m}CURE:}
Fig.~\ref{Fig4} and Tab.~\ref{tab1} illustrate that our method exhibits excellent performance across all sessions on the \emph{m}CURE dataset. In the base session, it achieves a notable $93.85\%$ accuracy, outperforming most non-compatibility methods and essentially on par with the best-performing method, SSFE-Net. Compared to other compatibility methods, our approach excels FACT by $9.85\%$, ALICE by $42.75\%$, and SAVC by $4.22\%$, demonstrating the efficacy of our forward-compatibility strategy. Thanks to our backward-compatibility approach, our method significantly outshines forward-compatibility focused methods in the final session, surpassing FACT by $20.84\%$, ALICE by $40.93\%$ and SAVC by $11.01\%$. Moreover, it outperforms non-compatibility methods in the final session, exceeding CEC by $19.75\%$, LIMIT by $19.84\%$, SSFE-Net by $5.44\%$ and BiDistFSCIL by $35.93\%$. Regarding AA, our framework substantially surpasses the best-performing compatibility and non-compatibility methods by $8.59\%$ and $3.15\%$, respectively. With a PD rate of only $15.69\%$, our method demonstrates effective resistance to forgetting among compatibility methods. This indicates that our approach extensively surpasses existing advanced methods on the \emph{m}CURE dataset.

\subsection{Ablation Studies}

To substantiate the significance of our proposed components, we conducted ablation studies focusing on the key aspects of our method, which consists of forward-compatible learning and backward-compatible learning. Forward-compatible learning encompasses virtual class generation and CT loss, while backward-compatible learning includes DR and KD based on PFS and uncertainty-guided selection. In Tab.~\ref{tab2}, we report the results starting with the CE loss-based fine-tuning as the baseline and progressively integrating virtual class generation, CT loss, raw PFS, and PFS with uncertainty quantification.

\begin{table*}[htbp!]
\centering
\renewcommand{\arraystretch}{1.2}
\setlength\tabcolsep{5.5pt}
\caption{Ablation studies on FCPill and \emph{m}CURE. VCG, CT, PFS, and US denote virtual class generation, CT loss, PFS, and uncertainty-guided selection, respectively. (In \%).} 
\begin{tabular}{cccccccccccccccc}
\hline
\multirow{2}{*}{Dataset}        & \multirow{2}{*}{VCG} & \multirow{2}{*}{CT loss} & \multirow{2}{*}{PFS} & \multirow{2}{*}{US} & \multicolumn{9}{c}{Session ID}                                                                                                                         & \multirow{2}{*}{AA$\uparrow$} & \multirow{2}{*}{PD$\downarrow$} \\ \cline{6-14}
                                &                      &                          &                      &                              & 0              & 1              & 2              & 3              & 4              & 5              & 6              & 7              & 8              &                     &                     \\ \hline
\multirow{5}{*}{FCPill}         &                      &                          &                      &                              & 95.01          & 86.87          & 80.04          & 74.33          & 69.58          & 65.50          & 58.18          & 54.27          & 51.52          & 70.59               & 43.49               \\
                                & \checkmark                    &                          &                      &                              & 96.18          & 88.65          & 81.06          & 77.53          & 73.98          & 69.83          & 64.83          & 63.55          & 61.98          & 75.29               & 34.20               \\
                                & \checkmark                    & \checkmark                        &                      &                              & 96.38          & 90.05          & 83.62          & 79.98          & 75.24          & 71.15          & 67.32          & 63.70          & 61.11          & 76.50               & 35.28               \\
                                & \checkmark                    & \checkmark                        & \checkmark                    &                              & 96.38          & 93.80          & 91.71          & 91.19          & 89.89          & 89.45          & 89.95          & 89.83          & 88.76          & 91.22               & 7.62                \\
                                & \checkmark                    & \checkmark                        & \checkmark                    & \checkmark                            & \textbf{96.38} & \textbf{94.54} & \textbf{92.74} & \textbf{92.03} & \textbf{91.04} & \textbf{90.41} & \textbf{90.68} & \textbf{90.66} & \textbf{89.59} & \textbf{92.01}      & \textbf{6.79}       \\ \hline
\multirow{5}{*}{\emph{m}CURE} &                      &                          &                      &                              & 84.47          & 72.57          & 64.35          & 58.64          & 53.52          & 48.76          & 46.10          & 40.18          & 37.66          & 56.25               & 46.82               \\
                                & \checkmark                    &                          &                      &                              & 91.00          & 77.68          & 64.71          & 60.86          & 55.04          & 51.29          & 45.87          & 42.52          & 41.49          & 58.94               & 49.50               \\
                                & \checkmark                    & \checkmark                        &                      &                              & 93.85          & 82.47          & 74.83          & 68.93          & 62.73          & 57.51          & 54.33          & 49.86          & 48.24          & 65.86               & 45.61               \\
                                & \checkmark                    & \checkmark                        & \checkmark                    &                              & 93.85          & 91.67          & 88.15          & 87.48          & 84.00          & 83.63          & 80.81          & 78.86          & 78.15          & 85.18               & 15.69               \\
                                & \checkmark                    & \checkmark                        & \checkmark                    & \checkmark                            & \textbf{93.85} & \textbf{91.67} & \textbf{88.15} & \textbf{87.48} & \textbf{84.00} & \textbf{83.63} & \textbf{80.81} & \textbf{78.86} & \textbf{78.15} & \textbf{85.18}      & \textbf{15.69}      \\ \hline
\end{tabular}
\label{tab2}
\end{table*}

On the FCPill dataset, compared to the baseline, the introduction of virtual class generation resulted in a 1.17\% improvement in the base session and a 10.46\% improvement in the final incremental session. The inclusion of CT loss further led to a 1.21\% improvement in AA. The raw PFS enhanced the performance by 14.72\% in AA. The integration of uncertainty-guided selection finally increased performance to 92.01\%, a 0.79\% improvement, in AA.

A similar trend was observed on the \emph{m}CURE dataset. Virtual class generation contributed to a 6.53\% improvement in the base session and a 3.83\% improvement in the last incremental session. The addition of CT loss led to further improvements of 2.85\% and 6.75\% in the base and final sessions, respectively. The raw PFS increased the AA performance by 19.32\%. Although the final integration of uncertainty-guided selection did not improve performance on the \emph{m}CURE, this indicates that the unfiltered pseudo-features were already capable of adequately simulating real features. Integrating the uncertainty filtering module can prevent additional risks due to poor synthetic pseudo-features.

\begin{figure}[htbp!]
    \centering
        \begin{subfigure}{0.32\linewidth}
        \centering
        \includegraphics[width=\linewidth]{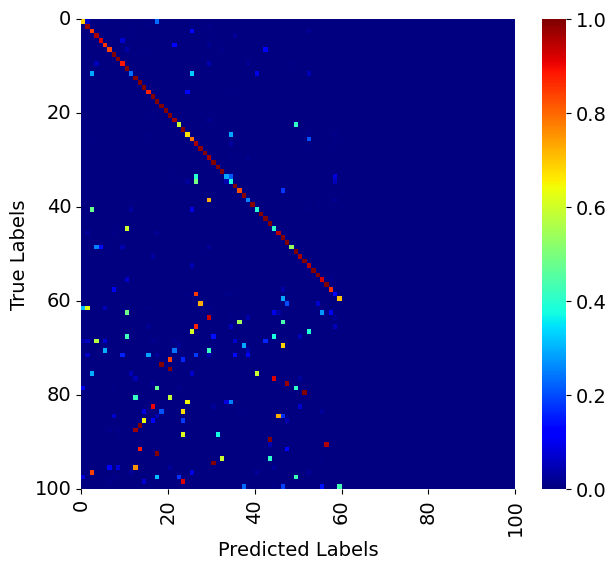}
        \caption{FT on FCPill}
    \end{subfigure}
    \begin{subfigure}{0.32\linewidth}
        \centering
        \includegraphics[width=\linewidth]{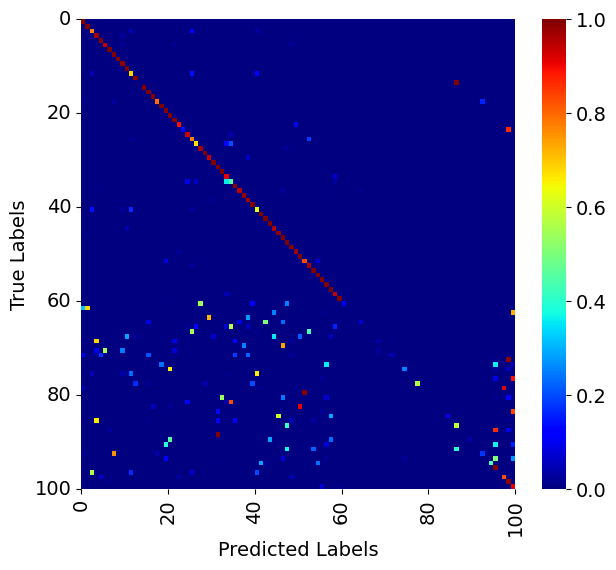}
        \caption{FCL on FCPill}
    \end{subfigure}    
        \begin{subfigure}{0.32\linewidth}
        \centering
        \includegraphics[width=\linewidth]{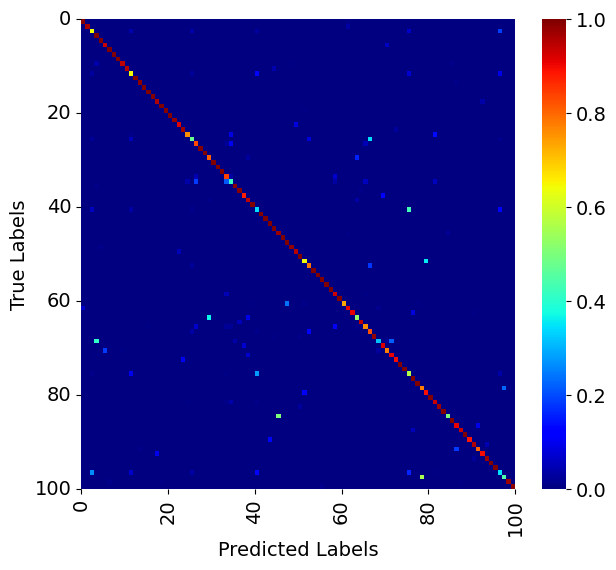}
        \caption{BCL on FCPill}
    \end{subfigure}    
        \begin{subfigure}{0.32\linewidth}
        \centering
        \includegraphics[width=\linewidth]{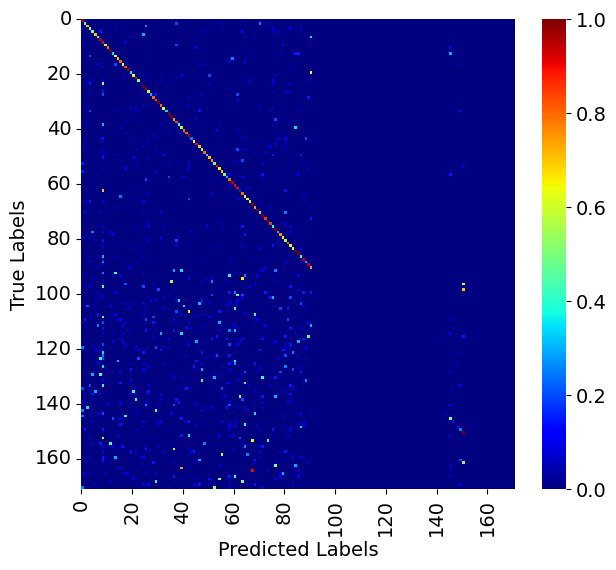}
        \caption{FT on \emph{m}CURE}
    \end{subfigure}
    \begin{subfigure}{0.32\linewidth}
        \centering
        \includegraphics[width=\linewidth]{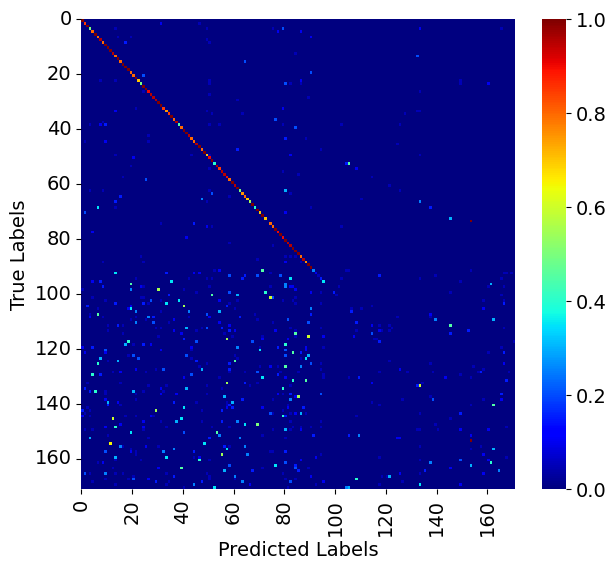}
        \caption{FCL on \emph{m}CURE}
    \end{subfigure}    
        \begin{subfigure}{0.32\linewidth}
        \centering
        \includegraphics[width=\linewidth]{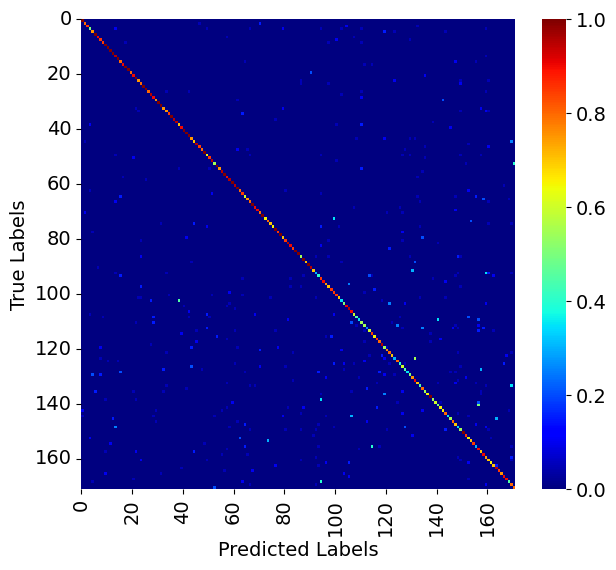}
        \caption{BCL on \emph{m}CURE}
    \end{subfigure}  
    \caption{Comparison of the confusion matrices of different ablation methods on FCPill and \emph{m}CURE datasets. FT, FCL, and BCL denote fine-tuning, forward-compatible learning (including VCG and CT loss), and back-compatible learning (including PFS and US).}
    \label{Fig5}
\end{figure}

To investigate the contributions of each component of our model in few-shot incremental pill recognition, we display the confusion matrices generated by models in the last incremental session of our ablation studies on the FCPill and \emph{m}CURE datasets in Fig.~\ref{Fig5}. A bright diagonal against a dim background indicates higher classification accuracy. Our observations reveal that while the basic fine-tuning method provides relatively clear diagonals for base classes, its effectiveness on new classes is limited. However, with the integration of different components, the model exhibits a noticeable improvement in performance for both new and old classes. Particularly, the implementation of forward-compatible learning significantly enhances performance for base classes while also improving results for incremental classes. Further incorporating the backward-compatible strategy markedly boosts performance for incremental classes. This demonstrates that our method effectively adapts to new classes and accurately recognizes old classes, avoiding confusion in established decision boundaries.

\subsection{Further Analysis}
\subsubsection{Effectiveness of Virtual Class Generation}
\begin{figure}[htbp!]
    \centering
        \begin{subfigure}{0.49\linewidth}
        \centering
        \includegraphics[width=\linewidth]{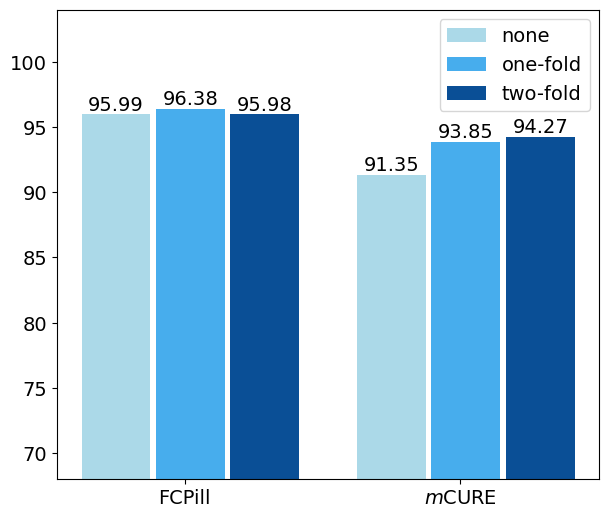}
        \caption{Accuracy on the base session}
    \end{subfigure}
    \hfill
    \begin{subfigure}{0.49\linewidth}
        \centering
        \includegraphics[width=\linewidth]{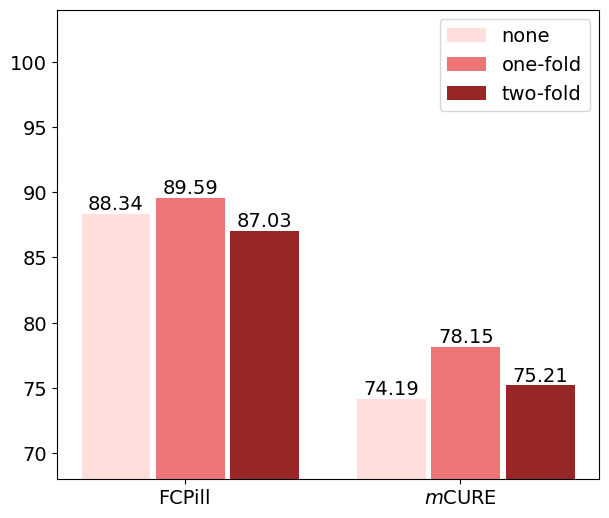}
        \caption{Accuracy on the final session}
    \end{subfigure}    
    \caption{Influence of virtual class generation methods. One-fold virtual class generation obtains the best performance.}
    \label{Fig6}
\end{figure}

To assess the efficacy of our proposed method for generating virtual classes, we present comparative performance results for both the base class session and the last incremental session on the FCPill and mCURE datasets, as shown in Fig.~\ref{Fig6}. We examine scenarios including no virtual class generation, one-fold virtual class generation, and two-fold virtual class generation. On the FCPill dataset, the one-fold virtual class generation strategy demonstrates superior performance over the no virtual class generation scenario, with improvements of 0.39\% and 1.25\% in the base class session and the last incremental session, respectively. This strategy also outperforms the two-fold virtual class generation, achieving improvements of 0.4\% and 2.56\% in the base class session and the last incremental session, respectively. Similarly, on the mCURE dataset, the one-fold virtual class generation approach shows enhanced results compared to the absence of virtual class generation, with performance boosts of 2.5\% and 3.96\% in the base class session and the last incremental session, respectively, and surpasses the two-fold virtual class generation with an improvement of 2.94\% in the last incremental session.

\subsubsection{Separation Capability of Center-Triplet Loss}
\begin{figure}[htbp!]
    \centering
        \begin{subfigure}{0.325\linewidth}
        \centering
        \includegraphics[width=\linewidth]{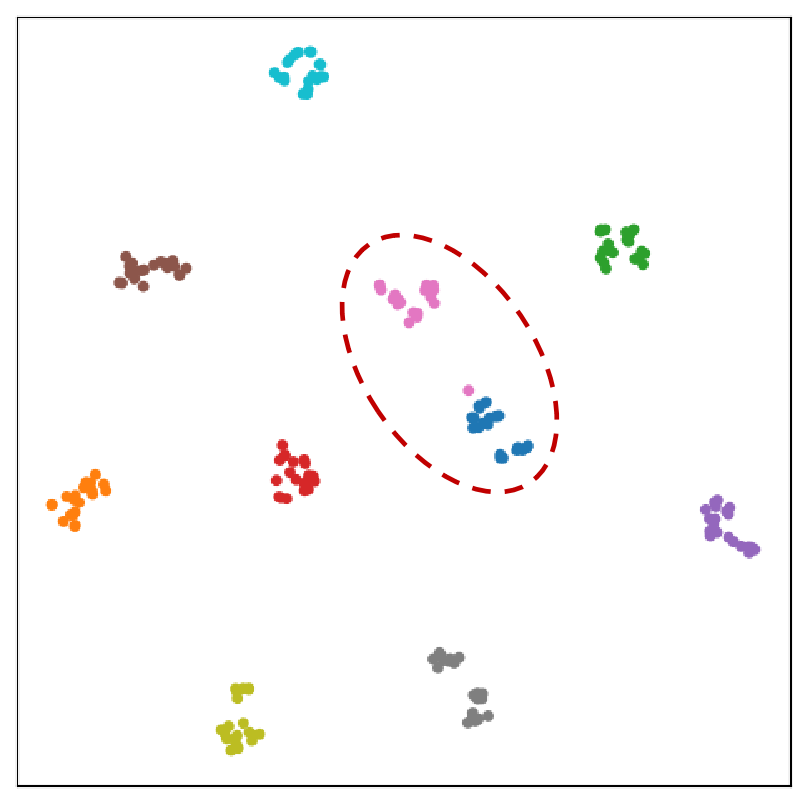}
        \caption{C. loss on FCPill}
    \end{subfigure}
    \begin{subfigure}{0.325\linewidth}
        \centering
        \includegraphics[width=\linewidth]{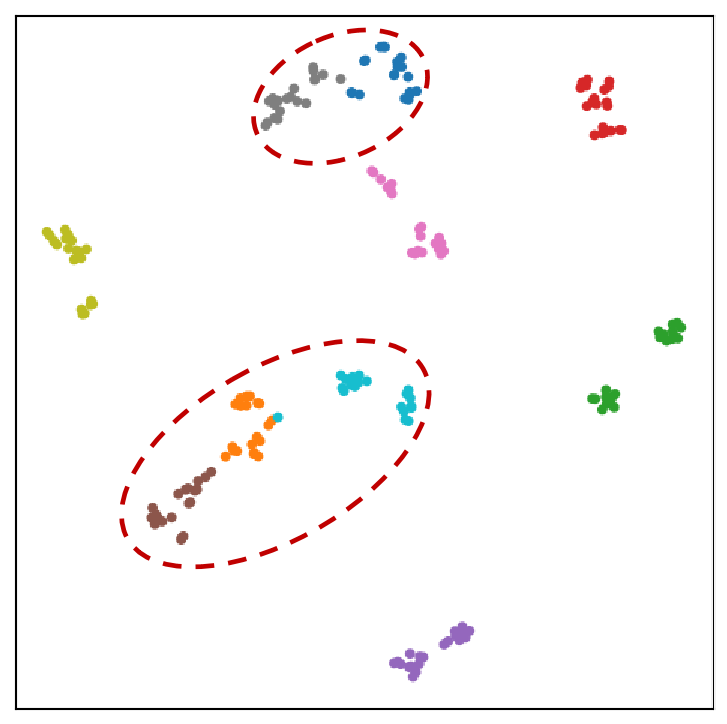}
        \caption{T. loss on FCPill}
    \end{subfigure}    
    \begin{subfigure}{0.325\linewidth}
        \centering
        \includegraphics[width=\linewidth]{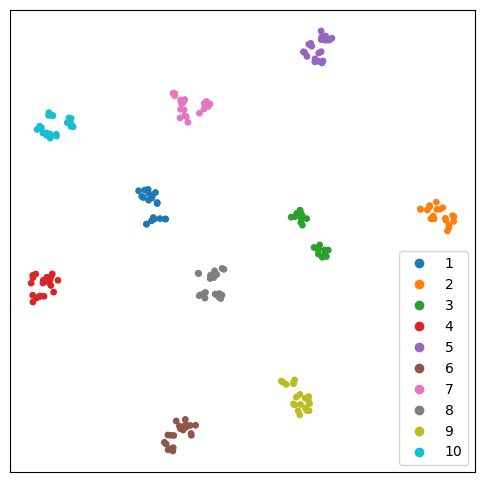}
        \caption{CT loss on FCPill}
    \end{subfigure}  
        \centering
        \begin{subfigure}{0.325\linewidth}
        \centering
        \includegraphics[width=\linewidth]{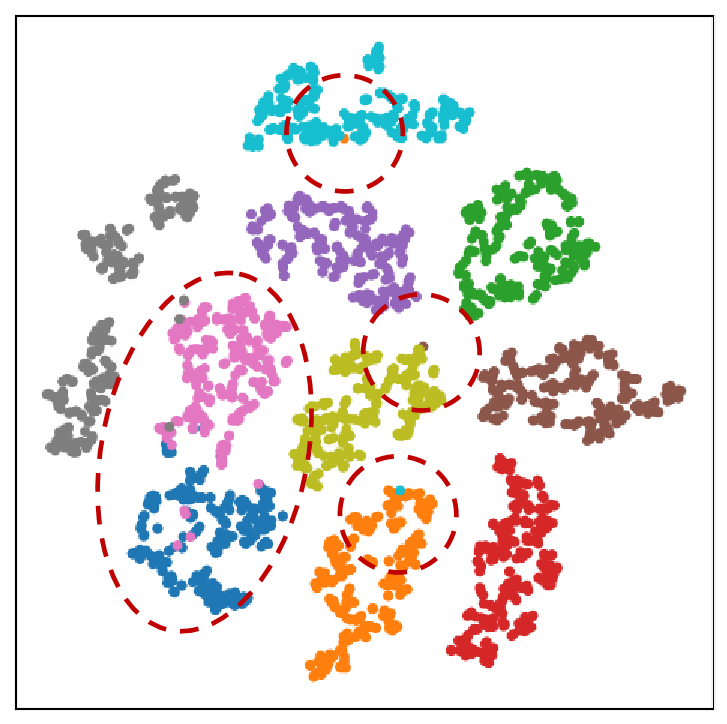}
        \caption{C. loss on \emph{m}CURE}
    \end{subfigure}
    \begin{subfigure}{0.325\linewidth}
        \centering
        \includegraphics[width=\linewidth]{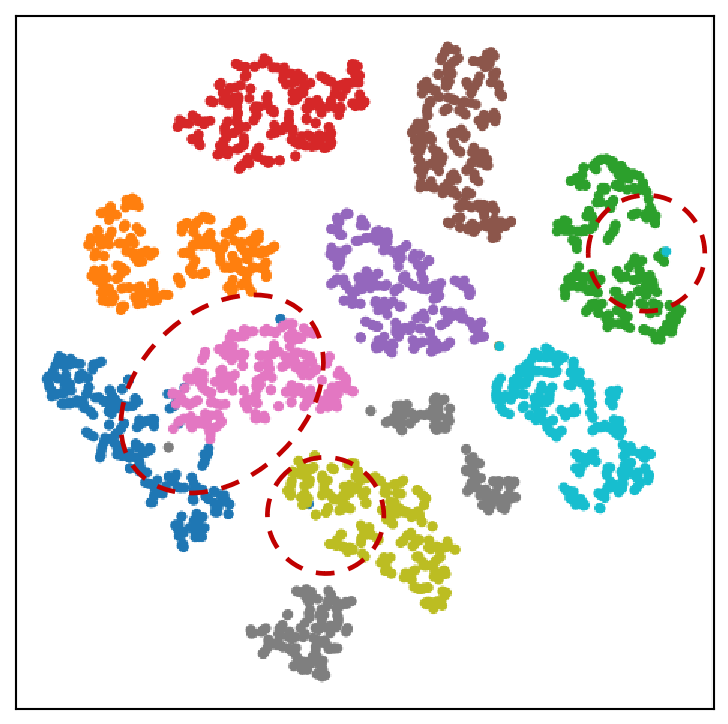}
        \caption{T. loss on \emph{m}CURE}
    \end{subfigure}    
    \begin{subfigure}{0.325\linewidth}
        \centering
        \includegraphics[width=\linewidth]{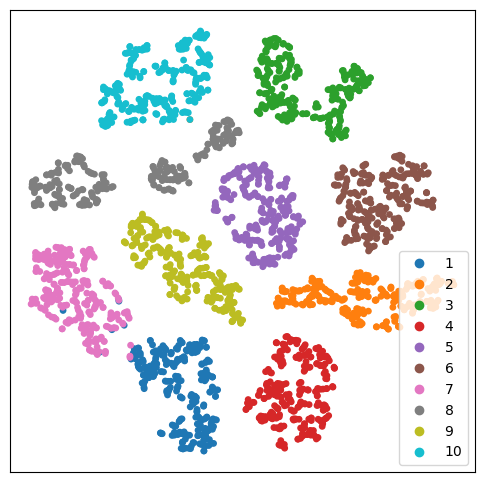}
        \caption{CT loss on \emph{m}CURE}
    \end{subfigure}  
    \caption{The t-SNE visualization of the features learned by different loss functions on FCPill and \emph{m}CURE datasets. Classes 0-6 represent the base classes, while classes 7-10 represent the incremental classes. Our CT loss gets the best class separation degree.}
    \label{Fig7}
\end{figure}

To verify the effectiveness of our CT loss in enhancing the separation of new and old classes, we visualized the feature space of the FCPill and \emph{m}CURE datasets in the last session using t-SNE, as depicted in Fig.~\ref{Fig7}. We randomly selected 6 base classes and 4 incremental classes, and compared the separation degree in the feature space under CE loss, Triplet Loss, Center Loss, and our proposed CT loss. The observations indicate that the performance of CE loss is the least effective; while Triplet Loss improves inter-class separation, it fails to constrain intra-class compactness. Although Center Loss can constrain intra-class compactness, it offers limited improvement in inter-class separation. Our proposed CT loss demonstrates the most significant effect in inter-class separation and intra-class compactness.

\subsubsection{Analysis of Pseudo Feature Synthesis}
To validate the effectiveness of our proposed PFS method, we visualized the feature space of the FCPill and \emph{m}CURE datasets in the final session using t-SNE, as shown in Fig.~\ref{Fig8}. We randomly selected some classes and displayed the distribution of real features and synthesized pseudo-features in the feature space. The observations reveal that the synthesized pseudo-features closely cluster with the real sample features on both FCPill and \emph{m}CURE datasets. This demonstrates that the synthesized pseudo-features effectively emulate the features generated by real samples, thereby aiding in the process of backward-compatible learning.

\begin{figure}[htbp!]
    \centering
        \begin{subfigure}{0.49\linewidth}
        \centering
        \includegraphics[width=\linewidth]{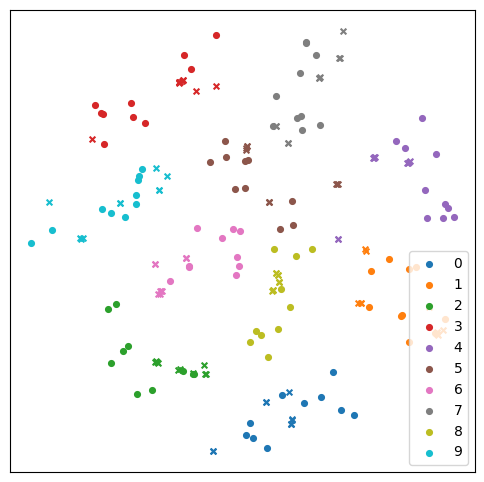}
        \caption{Features on FCPill}
    \end{subfigure}
    \begin{subfigure}{0.49\linewidth}
        \centering
        \includegraphics[width=\linewidth]{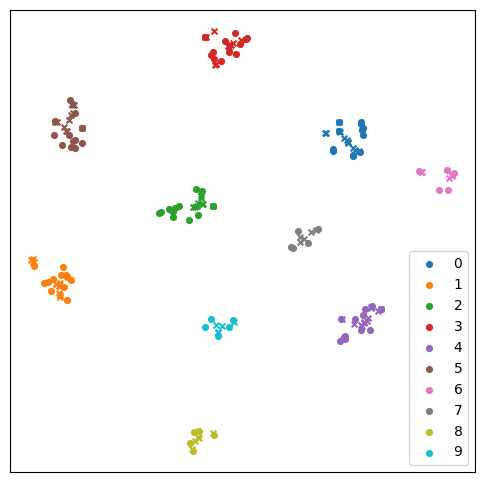}
        \caption{Features on \emph{m}CURE}
    \end{subfigure}    
    \caption{The t-SNE visualization of the real and pseudo features on FCPill and \emph{m}CURE datasets. The round and cross marks represent real and false features, respectively. Our PFS method effectively mimics the real features.}
    \label{Fig8}
\end{figure}

\subsubsection{Impact of Hyper-parameter}
In our DBC-FSCIL framework, the hyper-parameters for forward-compatible learning include the coefficient $\lambda$, as defined in Eq.~\ref{eq5}, which determines the impact of CT loss, and the margin $m$ in CT loss, as detailed in Eq.~\ref{eq4}. For backward-compatible learning, the hyper-parameters include $P$ and $Q$ in Alg.~\ref{alg1}, representing the number of real features stored per class and the number of pseudo-features synthesized per class, respectively, along with the temperature $T$ and the coefficient $\beta$ related to KD loss, as outlined in Eq.~\ref{eq9} and Eq.~\ref{eq10}. To thoroughly evaluate the Influence of these hyper-parameters on model performance, we present the results of the final session on the FCPill and \emph{m}CURE datasets with varying hyper-parameters in Fig.~\ref{Fig8}. It is observed that the optimal hyper-parameters for achieving the best performance on the FCPill dataset are $\{\lambda, m, P, Q, T, \beta\} = \{0.05, 1, 5, 10, 3, 0.4\}$. Similarly, for the \emph{m}CURE dataset, the hyper-parameters yielding the highest performance are $\{\lambda, m, P, Q, T, \beta\} = \{0.1, 2, 4, 12, 5, 0.6\}$.

\begin{figure}[htbp!]
    \centering
        \begin{subfigure}{0.326\linewidth}
        \centering
        \includegraphics[width=\linewidth]{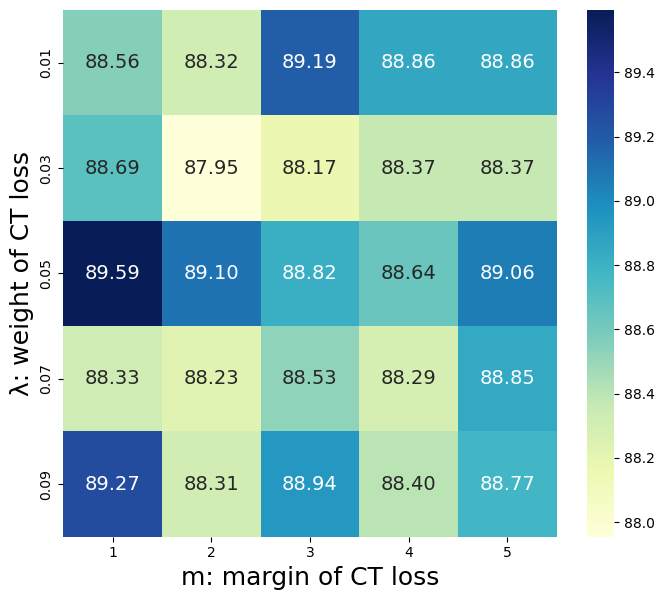}
        \caption{CT loss on FCPill}
    \end{subfigure}
        \begin{subfigure}{0.326\linewidth}
        \centering
        \includegraphics[width=\linewidth]{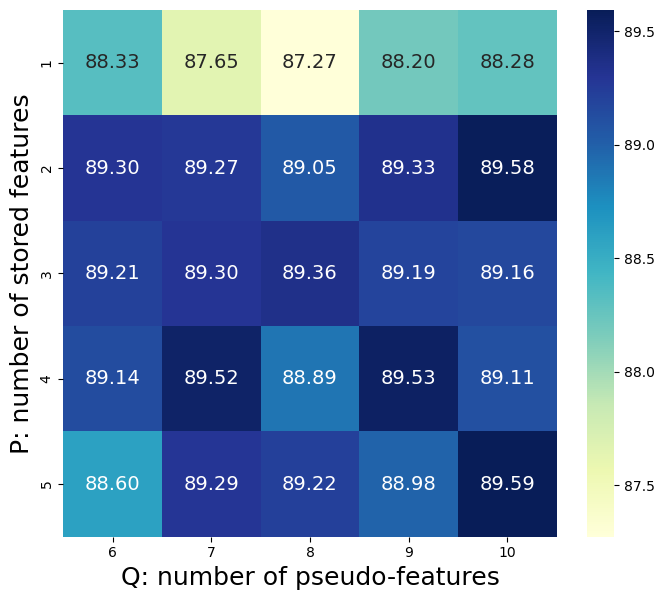}
        \caption{PFS on FCPill}
    \end{subfigure}
    \begin{subfigure}{0.326\linewidth}
        \centering
        \includegraphics[width=\linewidth]{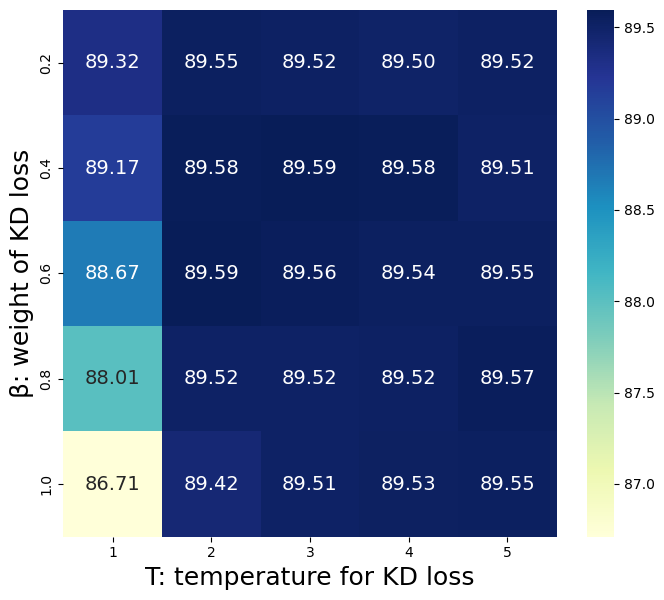}
        \caption{KD on FCPill}
    \end{subfigure}
    \begin{subfigure}{0.326\linewidth}
        \centering
        \includegraphics[width=\linewidth]{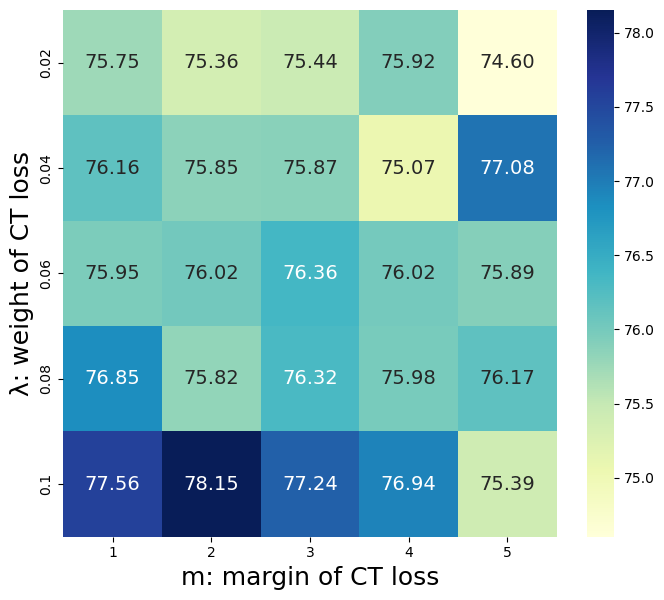}
        \caption{CT loss on \emph{m}CURE}
    \end{subfigure}
    \begin{subfigure}{0.326\linewidth}
        \centering
        \includegraphics[width=\linewidth]{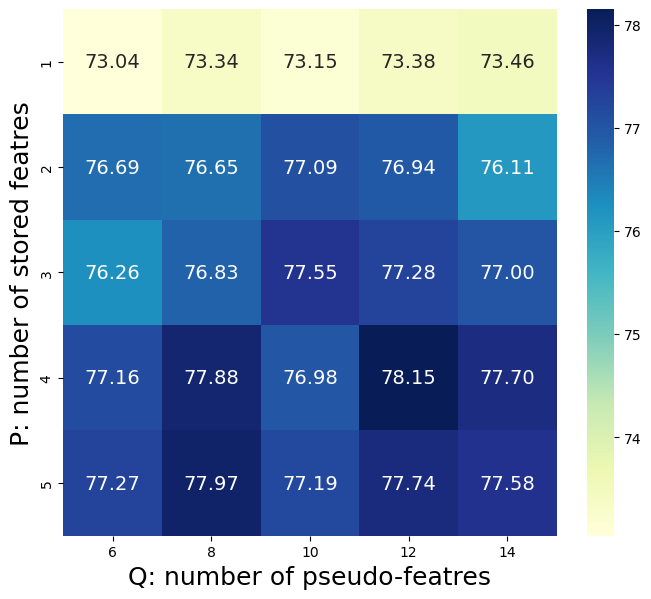}
        \caption{PFS on \emph{m}CURE}
    \end{subfigure}
    \begin{subfigure}{0.326\linewidth}
        \centering
        \includegraphics[width=\linewidth]{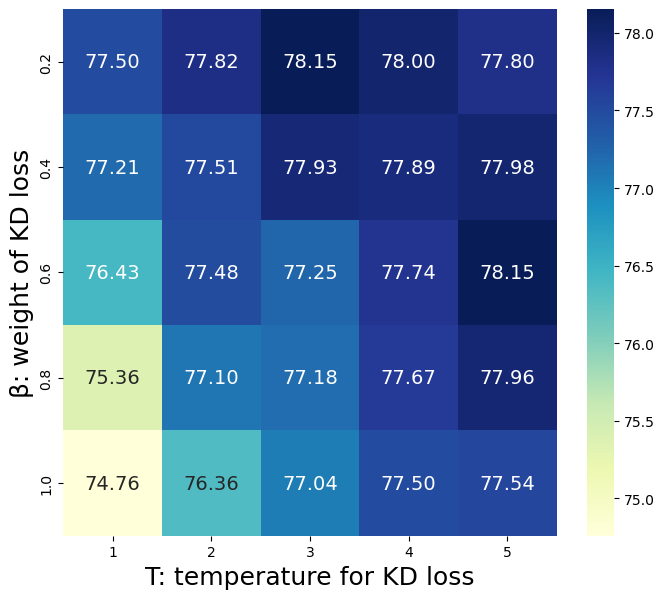}
        \caption{KD on \emph{m}CURE}
    \end{subfigure}    
    \caption{Hyper-parameter influence on the FCPill and \emph{m}CURE datasets.}
    \label{Fig8}
\end{figure}

\section{Conclusion}
\label{Conclusion}

In this paper, we introduce the first FSCIL framework for pill recognition, named DBC-FSCIL. This framework incorporates forward-compatible and backward-compatible learning components. For forward-compatible learning, we propose an innovative virtual class synthesis strategy and a CT loss to enhance discriminative feature learning. These virtual classes act as placeholders in the feature space for future class updates, providing diverse semantic knowledge for model training. Regarding backward-compatible learning, we develop a strategy to synthesize reliable pseudo-features of old classes using uncertainty quantification, facilitating DR and KD. This approach enables flexible feature synthesis and significantly reduces the additional storage requirements for samples and models. Furthermore, we have constructed a new pill image dataset for FSCIL and assessed various mainstream FSCIL methods, establishing new benchmarks. Our experimental results demonstrate that our framework surpasses existing SOTA methods.

\footnotesize
\bibliographystyle{IEEEtran}
\bibliography{IEEEabrv,myreference}

\begin{IEEEbiography}[{\includegraphics[width=1in,height=1.25in,clip,keepaspectratio]{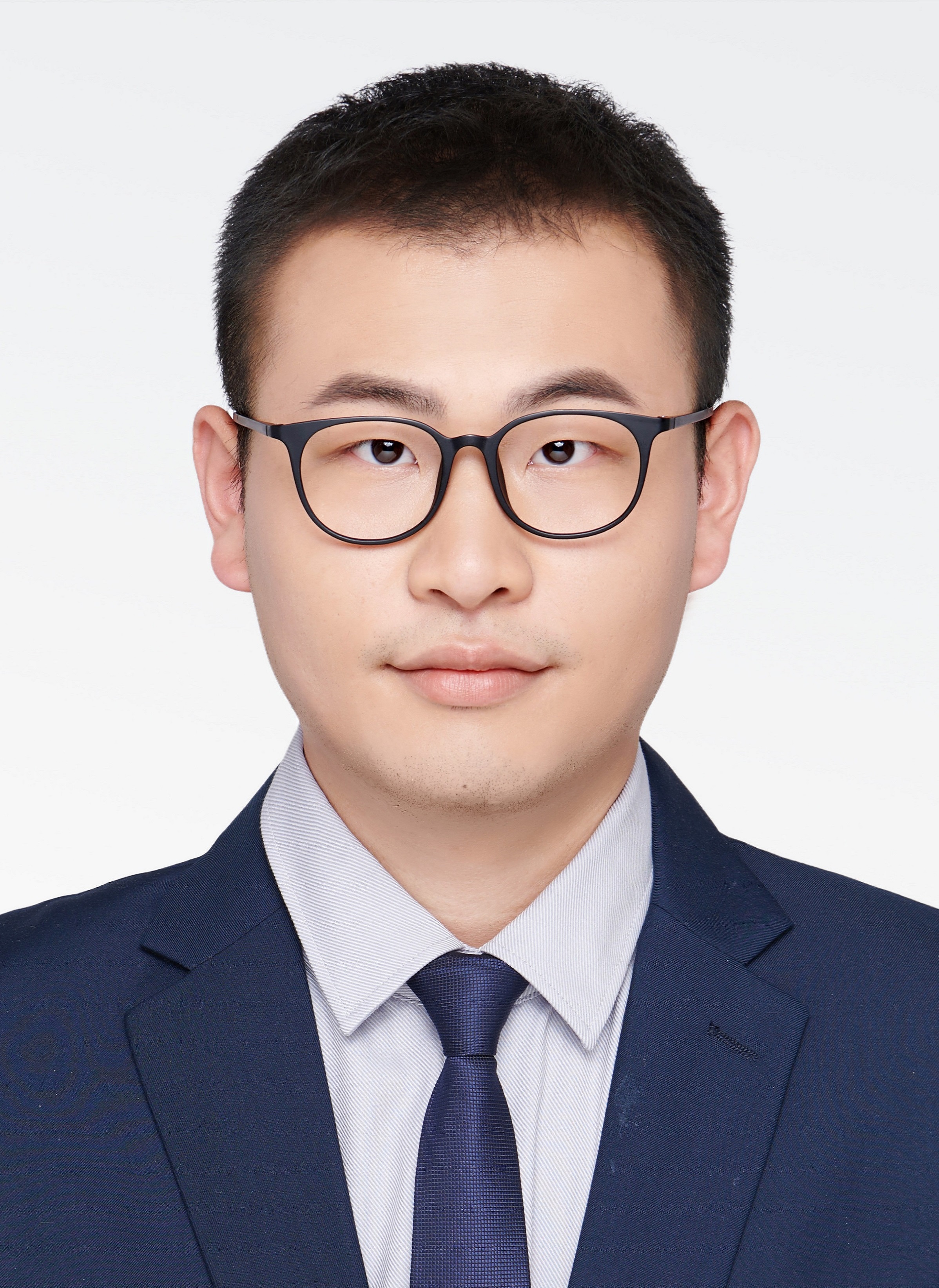}}]{Jinghua Zhang} received the B.E. degree from the Hefei University, China, in 2018, and the M.E. degree from the Northeastern University, China, in 2021. He is currently pursuing the Ph.D. degree in control science and engineering with the National University of Defense Technology, and is also with the Center for Machine Vision and Signal Analysis, University of Oulu. His research interests include computer vision and deep learning.
\end{IEEEbiography}

\begin{IEEEbiography}[{\includegraphics[width=1in,height=1.25in,clip,keepaspectratio]{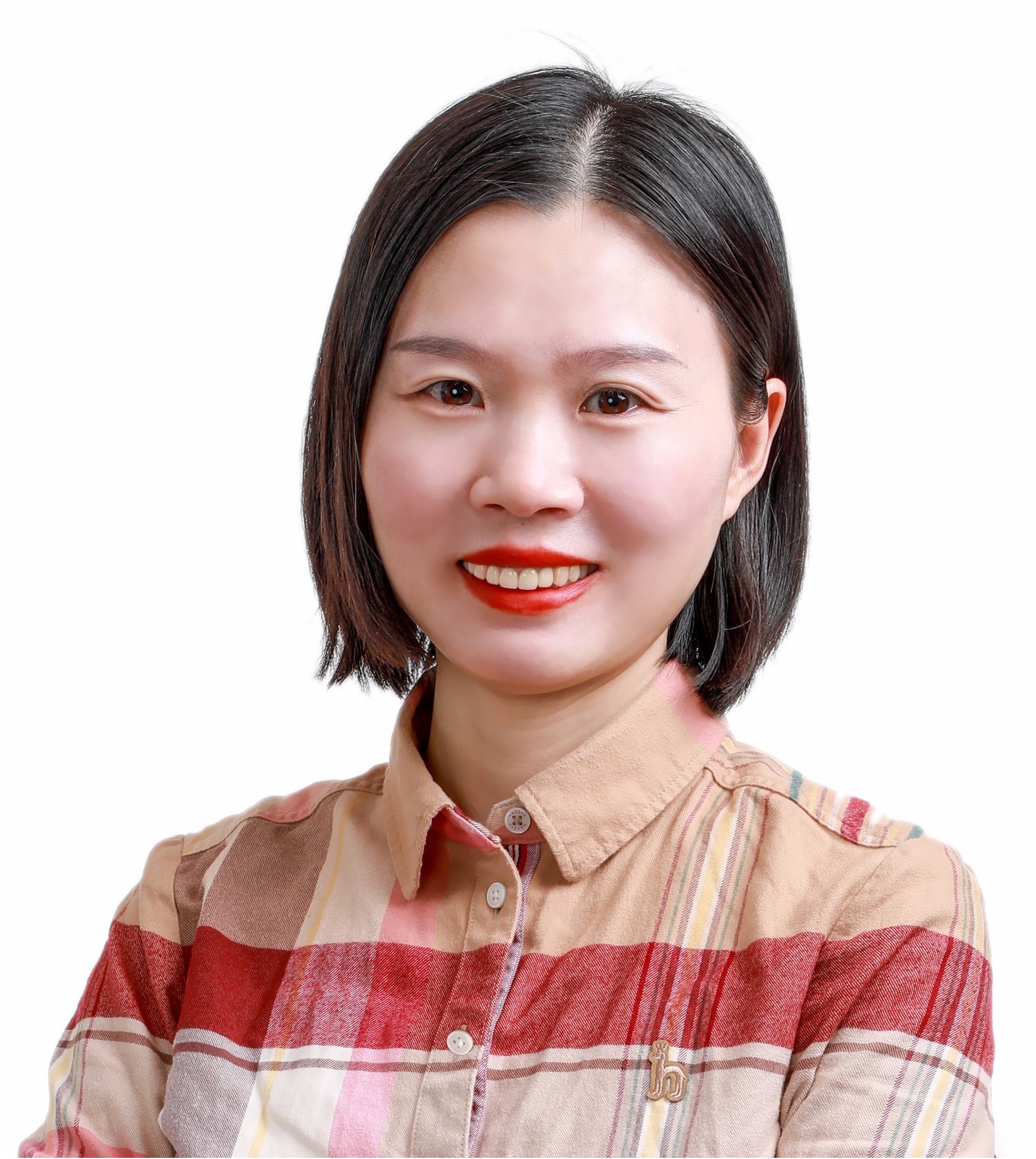}}]{Li Liu} received the Ph.D. degree in information and communication engineering from the National University of Defense Technology (NUDT), Changsha, China, in 2012.,She is currently a Full Professor with NUDT. She has held visiting appointments with the University of Waterloo in Canada, at the Chinese University of Hong Kong, and at the University of Oulu in Finland. Her research interests include computer vision, pattern recognition, and machine learning.,Dr. Liu served as a co-chair of many International Workshops along with major venues like CVPR and ICCV. She served as the leading guest editor of the special issues for IEEE TPAMI and IJCV. She also served as an Area Chair for several respected international conferences. She currently serves as an Associate Editor for IEEE TCSVT and Pattern Recognition. Her papers currently have over 13 000 citations, according to Google Scholar.
\end{IEEEbiography}

\begin{IEEEbiography}[{\includegraphics[width=1in,height=1.25in,clip,keepaspectratio]{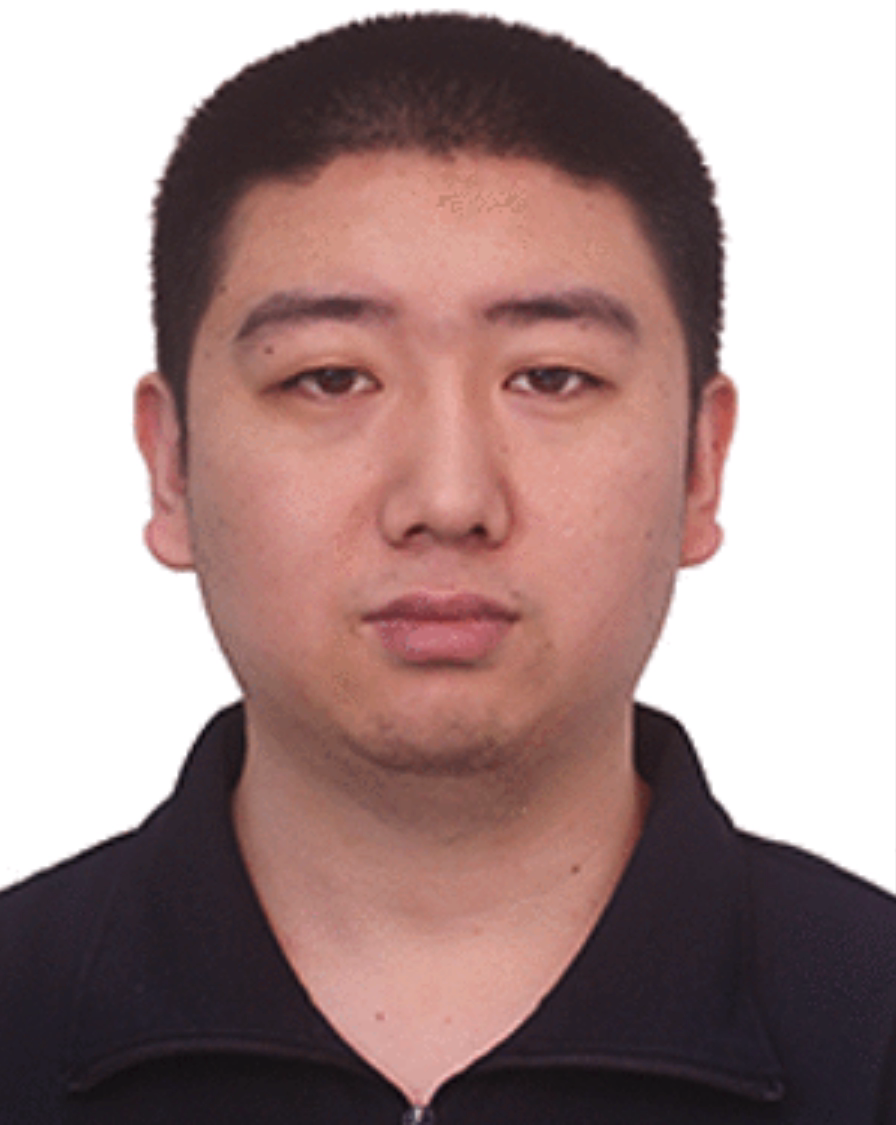}}]{Kai Gao} received the BSc, MSc, and PhD degrees from the National University of Defense Technology, China, in 2014, 2017, and 2023, respectively. His research interests include medical image analysis, pattern recognition, and machine learning.
\end{IEEEbiography}

\begin{IEEEbiography}[{\includegraphics[width=1in,height=1.25in,clip,keepaspectratio]{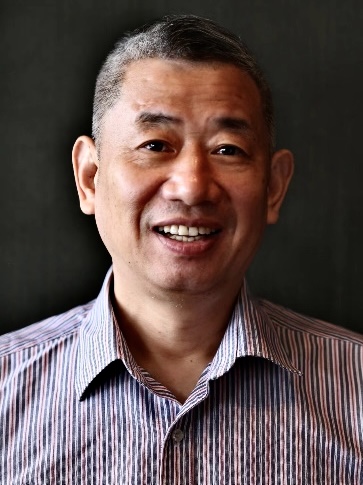}}]{Dewen Hu} received the B.S. and M.S. degrees from Xi’an Jiaotong University, Xi’an, China, in 1983 and 1986, respectively, and the Ph.D. degree from the National University of Defense Technology, Changsha, China, in 1999.,From 1995 to 1996, he was a Visiting Scholar with the University of Sheffield, Sheffield, U.K. He is currently a Professor with the School of Intelligent Science, National University of Defense Technology. His research interests include image processing, system identification and control, neural networks, and cognitive science.
\end{IEEEbiography}

 \newpage
\end{document}